\documentclass{article}
\usepackage{arxiv}
\usepackage[utf8]{inputenc}
\usepackage[T1]{fontenc}
\usepackage{hyperref}
\usepackage{url}
\usepackage{booktabs}
\usepackage{amsfonts}
\usepackage{nicefrac}
\usepackage{microtype}
\usepackage{graphicx}
\usepackage{natbib}
\usepackage{doi}
\usepackage{multirow}
\usepackage{amsmath} 

\title{Beyond Vulnerabilities: A Survey of Adversarial Attacks as Both Threats and Defenses in Computer Vision Systems}

\author{
    Zhongliang Guo
    \And
    Yifei Qian
    \And
    Yanli Li
    \And
    Weiye Li
    \And
    Chun Tong Lei
    \And
    Shuai Zhao
    \And
    Lei Fang
    \And
    Ognjen Arandjelovi\'c
    \And
    Chun Pong Lau
}

\date{}
\renewcommand{\undertitle}{}
\renewcommand{\shorttitle}{}

\begin{document}
\maketitle

\begin{abstract}
Adversarial attacks against computer vision systems have emerged as a critical research area that challenges the fundamental assumptions about neural network robustness and security. This comprehensive survey examines the evolving landscape of adversarial techniques, revealing their dual nature as both sophisticated security threats and valuable defensive tools. We provide a systematic analysis of adversarial attack methodologies across three primary domains: pixel-space attacks, physically realizable attacks, and latent-space attacks. Our investigation traces the technical evolution from early gradient-based methods such as FGSM and PGD to sophisticated optimization techniques incorporating momentum, adaptive step sizes, and advanced transferability mechanisms. We examine how physically realizable attacks have successfully bridged the gap between digital vulnerabilities and real-world threats through adversarial patches, 3D textures, and dynamic optical perturbations. Additionally, we explore the emergence of latent-space attacks that leverage semantic structure in internal representations to create more transferable and meaningful adversarial examples. Beyond traditional offensive applications, we investigate the constructive use of adversarial techniques for vulnerability assessment in biometric authentication systems and protection against malicious generative models. Our analysis reveals critical research gaps, particularly in neural style transfer protection and computational efficiency requirements. This survey contributes a comprehensive taxonomy, evolution analysis, and identification of future research directions, aiming to advance understanding of adversarial vulnerabilities and inform the development of more robust and trustworthy computer vision systems.
\end{abstract}
\section{Introduction}
The remarkable advancement of deep neural networks (DNNs) has revolutionized artificial intelligence applications across numerous domains, with computer vision systems achieving unprecedented performance levels that often match or exceed human capabilities~\cite{jia2025uni,jia2025towards,guo2023siamese}. This technological evolution has led to widespread deployment of vision-based AI systems in critical sectors including autonomous driving, medical diagnosis, financial services, and security applications. However, alongside these achievements, fundamental vulnerabilities have emerged that challenge the reliability and trustworthiness of these systems.

\begin{figure}[t]
    \centering
    \includegraphics[width=1.0\columnwidth]{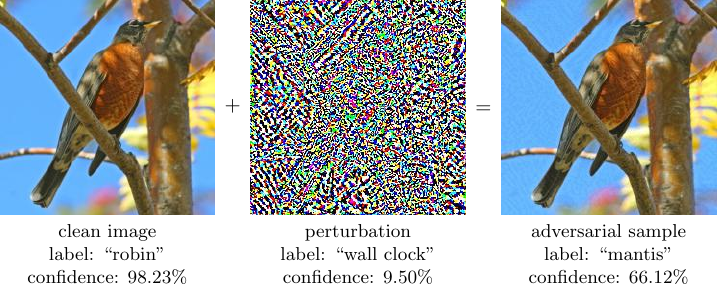}
    \vspace{-2em}
    \caption[Visualization of an adversarial sample making the classifier misclassification]{An exemplar for a successful adversarial attack. The attack is implemented by the Fast Gradient Sign Method~\cite{goodfellow2014explaining} on VGG-19~\cite{simonyan2014very} with eps=4/255. The clean image is correctly predicted by a classification model as robin with 98.23\% confidence. However, when We add perturbations that is not up to 4/255 for each pixel, although perturbations are imperceptible for human, the classification model predicts the adversarial sample as a wrong category.}
    \label{fig:intro:advexample}
\end{figure}

A pivotal discovery by Szegedy et al.~\cite{szegedy2013intriguing} revealed that neural networks are susceptible to adversarial attacks—carefully crafted inputs that can cause models to produce incorrect outputs while remaining imperceptible to human observers, as shown in Figure~\ref{fig:intro:advexample}. This vulnerability represents a significant security concern, as minor pixel-level perturbations can systematically fool state-of-the-art computer vision models. The implications extend far beyond academic curiosity, manifesting as real-world threats in safety-critical applications where model failures can have serious consequences.

Simultaneously, the rapid development of generative AI technologies has introduced new ethical and security challenges. The emergence of sophisticated generative models capable of producing highly realistic synthetic content has enabled both beneficial applications and malicious uses, including deepfakes~\cite{verdoliva2020media}, unauthorized content generation~\cite{guo2024grey,guo2025building}, and intellectual property violations~\cite{guo2024artwork,guo2025artwork}. This dual nature of AI technology—as both a powerful tool and a potential security risk—underscores the critical need for comprehensive understanding of adversarial vulnerabilities and protection mechanisms.

The implications of these vulnerabilities extend far beyond academic concerns, manifesting across critical real-world applications. In autonomous driving systems, adversarial patches on traffic signs can cause misclassification, potentially leading to safety hazards~\cite{eykholt2018robust}. Medical imaging systems face similar risks, where imperceptible perturbations to X-rays or CT scans could result in misdiagnosis~\cite{finlayson2018adversarial}. Financial institutions employing ML for fraud detection are vulnerable to adversarial examples that could allow fraudulent transactions to bypass security measures~\cite{cartella2021adversarial}. Content moderation systems on social media platforms can be fooled by adversarial examples, enabling the dissemination of harmful content~\cite{zhang2019adversarial}. Industrial quality control systems represent another critical vulnerability, where adversarial attacks could cause defective products to pass visual inspection~\cite{yuan2019adversarial}. Conversely, adversarial techniques have shown promise in beneficial applications, such as protecting individual privacy from unauthorized facial recognition systems~\cite{shan2020fawkes} and safeguarding intellectual property of trained models from unauthorized extraction~\cite{adi2018turning}. These diverse applications underscore the critical importance of understanding and addressing adversarial vulnerabilities in modern ML systems.

\subsection{Research Scope and Objectives}
This literature review provides a comprehensive examination of adversarial attacks in computer vision systems, focusing on three primary research dimensions:
\begin{enumerate}
    \item \textbf{Attack Methodologies:} A systematic analysis of adversarial attack techniques, including their underlying mechanisms, effectiveness, and evolution from basic perturbation methods to sophisticated attack strategies.
    \item \textbf{Vulnerability Assessment:} An exploration of how adversarial techniques serve as tools for evaluating the robustness boundaries of computer vision models, particularly in critical applications such as biometric authentication systems.
    \item \textbf{Defensive Applications:} An investigation into the constructive use of adversarial techniques for preventing malicious applications, including protection against unauthorized neural style transfer and misuse of generative models.
\end{enumerate}

The scope of this review encompasses pixel-space attacks, physically realizable attacks, and latent-space attacks, while also examining emerging applications in biometric security and generative model protection. We aim to provide researchers and practitioners with a comprehensive understanding of the current state of adversarial machine learning in computer vision and identify key research gaps and future directions.

\subsection{Contribution and Organization}
This review makes several key contributions to the field:
\begin{itemize}
    \item \textbf{Comprehensive Taxonomy:} We present a systematic categorization of adversarial attacks based on attacker knowledge, underlying mechanisms, objectives, and operational domains.
    \item \textbf{Evolution Analysis:} We trace the technical evolution from early gradient-based methods to sophisticated optimization techniques and their practical implications.
    \item \textbf{Application-Focused Perspective:} We examine adversarial techniques from both offensive and defensive standpoints, highlighting their dual role in vulnerability assessment and protection mechanisms.
    \item \textbf{Critical Gap Identification:} We identify underexplored research areas and emerging challenges that warrant future investigation.
\end{itemize}

The remainder of this review is organized as follows: Section 2 establishes fundamental concepts and evaluation frameworks for adversarial attacks. Sections 3-5 provide detailed examinations of pixel-space attacks, physically realizable attacks, and latent-space attacks respectively. Section 6 explores emerging applications in biometric security and generative model protection. Finally, Section 7 discusses future research directions and concludes the review.

Through this comprehensive analysis, we aim to advance understanding of adversarial vulnerabilities in computer vision systems and inform the development of more robust and trustworthy AI technologies.
\section{Preliminary}
This section provides an elementary introduction to adversarial attacks in computer vision systems. We begin by exploring the fundamental concepts of adversarial attacks, presenting a systematic categorization based on attacker knowledge, underlying mechanisms, objectives, and domains of operation. We also discuss the evaluation metrics used to assess the effectiveness of these attacks, including performance, transferability, and imperceptibility.

\subsection{Concept of Adversarial Attacks}

This paper start with the concept of adversarial attack. Given an image $\boldsymbol{x}\in \mathbb{R}^{ C \times H \times W}$ from a dataset $\mathcal{X}$, where $C$, $H$, $W$ is the number of color channels, the image height, and its width, respectively. An adversarial example $\boldsymbol{x}^{adv}$ is generated as follows:
\begin{equation}
    \boldsymbol{x}^{adv} = \boldsymbol{x} + \boldsymbol{\delta},
\end{equation}
where $\boldsymbol{\delta}$ refers to the adversarial perturbation. Here We unify any attack method as $A(\cdot)$, having $\boldsymbol{\delta} = A(\boldsymbol{x})$. The NN, denoted as $\mathcal{F}(\cdot)$, serves as a differentiable model that, when given an input, produces an output reflective of the specific task it is designed to perform. The adversarial perturbations will significantly deviate the output of a NN from its expected response. If we leverage $\mathcal{D}(\cdot)$ to measure the difference between pre-attack results and post-attack results, it can be denoted as $\max (\mathcal{D}(\mathcal{F}(\boldsymbol{x}),\mathcal{F}(\boldsymbol{x}^{adv})))$.

In the majority of attack scenarios, adversarial perturbations are confined within a specific range $\epsilon$ to ensure they remain visually imperceptible~\cite{goodfellow2014explaining,madry2018towards}.
The measure of range is various, may include quantitatively controlling the size of the perturbation, or controlling the human perception changes towards the image.
This constraint allows the perturbations to subtly manipulate the NN's output without raising suspicion or being detectable by human observers.

Thus, an adversarial attack method can be formulated as:
\begin{equation}
    \arg \max_{\left\|\boldsymbol{\delta}\right\|\leq \epsilon} \mathcal{D}\big(\mathcal{F}(\boldsymbol{x}),\mathcal{F}(\boldsymbol{x}^{adv})\big)\,,
\end{equation}
where $\left\|\cdot\right\|$ denotes a kind of constraint, to maintain the visual integrity of adversarial samples.
In the practical execution of adversarial attacks,  The constraint $\left\|\cdot\right\|$ can represent various forms of limitations, such as the maximum allowable perturbation on individual pixels~\cite{goodfellow2014explaining}, ensuring minimal visual perceptibility~\cite{yuan2022natural}, or maintaining the semantic content of the image~\cite{liu2023instruct2attack}. This formulation encapsulates the essence of crafting adversarial examples that are effective in deceiving the model while being subtle enough to evade detection.

\subsection{Categories of Adversarial Attacks}
Adversarial attacks can be systematically categorized according to several fundamental criteria~\cite{xu2020adversarial,qiu2019review}. In this section, We will elucidate the mainstream taxonomies used to classify these attacks, including the attacker's knowledge of the target model, the underlying mechanisms employed, the specific objectives pursued, and the domains in which these attacks operate.

\subsubsection{Knowledge of Attackers}
Regarding to the attacker's knowledge about the target model, i.e., if the attacker can access the NN architecture and weights, the terms ``black-box'' and ``white-box'' refer to two distinct scenarios.
\paragraph{White-box attack.} 
White-box attack is one where the attacker possesses complete knowledge of the NN's architecture and parameters.
This comprehensive insight allows the attacker to craft adversarial examples with precision, exploiting the specific vulnerabilities of the model.
The attacker can utilize gradient-based methods~\cite{goodfellow2014explaining,madry2018towards} or other optimization techniques~\cite{dong2018boosting,zhang2023boosting} to generate inputs that are designed to lead the model astray, making white-box attacks particularly potent and challenging to defend against.

\paragraph{Grey-box attack.} 
Grey-box attack represents an intermediate scenario where the attacker possesses partial knowledge about the target model, such as the architecture but not the parameters, or access to similar training data without knowing the exact model weights. This limited information allows attackers to craft more effective adversarial examples compared to pure black-box scenarios by leveraging architectural similarities or training surrogate models with known components~\cite{papernot2017practical,tramer2018ensemble}. While less powerful than white-box attacks, grey-box approaches can significantly improve attack success rates by exploiting available structural knowledge about the target model.

\paragraph{Black-box attack.}
Black-box attack depicts a situation where the attacker has no direct access to the model's internals.
The attacker's knowledge is limited to the input data and the corresponding outputs from the model.
Despite this lack of detailed information, attackers can still craft effective adversarial examples.
They often employ techniques like transferability of adversarial examples from substitute models~\cite{dong2023restricted}, query-based methods~\cite{andriushchenko2020square}, or gradient estimation strategies~\cite{zhang2021progressive}.
\subsubsection{Underlying Mechanisms}
Adversarial attack techniques can be categorized based on their underlying mechanisms, broadly falling into two distinct types: gradient-based or optimization-based, and generative model-based approaches. Each of these offers different advantages and operates under varying assumptions about the attacker's knowledge and capabilities, contributing to the diverse and complex landscape of adversarial attacks in NNs.
\paragraph{Gradient-based (optimization-based) method.}
These methods leverage the gradient of the NN to identify the most effective alterations to the input data~\cite{goodfellow2014explaining}. Essentially, this kind of approach is to optimize a single image or adversarial perturbation to make the final output deceptive. By understanding how small changes in the input can significantly impact the output, these methods efficiently generate adversarial examples that are tailored to disrupt the network.
\paragraph{Generative model-based method.}
Generative model-based attack employs generative models like Generative Adversarial Networks (GANs), Autoencoders (AEs), and Diffusion Models (DMs) to craft adversarial examples~\cite{wang2023lfaa}. These models are trained to generate inputs that are perceptually indistinguishable from real data but are structured in a way that leads the victim network to make incorrect predictions. The essence of this method is to use NNs to learn a distribution in which the images in this distribution are deceptive to the victim model.
\subsubsection{Attack Objectives}
In the exploration of adversarial strategies, a crucial distinction lies in the objectives of the attacks, specifically differentiating between targeted and untargeted attacks. This differentiation is not limited to classification models or solely to the realm of computer vision but extends across various domains and tasks within computer vision.
\paragraph{Targeted Attacks.}
the adversary's goal of targeted attacks is to manipulate the model into producing a specific, incorrect output. The attacker aims not just to cause a misclassification or error but to steer the model toward a predetermined, incorrect result. This type of attack requires a more nuanced understanding of the model's behavior and often involves sophisticated perturbation techniques. In contexts beyond classification, such as object detection or segmentation in computer vision, a targeted attack might aim to make the model identify an object as a particular class erroneously or modify the boundaries of segmentation to match a specific, incorrect shape.
\paragraph{Untargeted Attacks.}
Contrastingly, untargeted attacks focus on the broader objective of simply causing the model to err, without any preference for the specific type of mistake. The primary aim is to degrade the overall performance of the model, ensuring that the output deviates from the correct or expected result, regardless of the specific incorrect outcome. In broader applications, such as in natural language processing or generative models, an untargeted attack could aim to disrupt the coherence of generated text or the fidelity of synthesized images without guiding the model toward any particular error.
\subsubsection{Attack Domain}
Adversarial attacks can be categorized based on the domain in which they operate, reflecting different approaches to manipulating inputs and exploiting model vulnerabilities.

\paragraph{Pixel-Space Attacks.}
Pixel-Space Attacks (PSAs) directly modify the raw input images at the pixel level to generate adversarial examples~\cite{goodfellow2014explaining,madry2018towards}. These attacks operate on the most immediate and explicit representation of visual data, making precise alterations to individual pixel values to induce misclassification. While the modifications are often imperceptible to human observers, they can significantly impact model predictions by exploiting the sensitivity of neural networks to small input changes. PSAs typically leverage the gradient-based / optimization-based framework, though there are few methods~\cite{liang2022lp,wang2023lfaa} used generative models to generate the pixel-space perturbations. PSAs often use $\ell_p$-norms as constrains respect to perturbations, ensuring the imperceptibility of adversarial samples.
        
\paragraph{Physically Realizable Attacks.}
Physically realizable attacks bridge the gap between digital perturbations and real-world implementations, addressing the challenges of creating adversarial examples that remain effective when captured through sensors or cameras~\cite{eykholt2018robust}. These attacks consider physical constraints such as lighting conditions, viewing angles, printing limitations, and environmental factors. Examples include adversarial patches, which can be printed and attached to objects; adversarial eyeglass frames that fool facial recognition systems; or adversarial textures applied to 3D objects. The defining characteristic of these attacks is their robustness against physical transformations and their ability to impact systems in practical deployment scenarios.

\paragraph{Latent-Space Attacks.}
Latent-Space Attacks (LSAs) operate by manipulating the intermediate representations or embeddings that deep learning models construct internally~\cite{liu2023instruct2attack}. Rather than modifying raw inputs, these attacks target the higher-dimensional feature spaces where semantic information is encoded. By perturbing these latent representations, attackers can achieve more efficient and semantically meaningful adversarial examples, often with greater transferability across different models. Such attacks leverage the understanding that models organize information in their hidden layers according to learned patterns and concepts, making these internal representations particularly vulnerable.

\subsection{Evaluation Protocol for Adversarial Attacks}
The evaluation protocol for adversarial attack can be divided into three aspects, performance, transferability, and imperceptibility.
The challenge in adversarial attack design lies in balancing attack effectiveness with imperceptibility, as these goals often conflict: stronger perturbations typically increase attack success rates but become more visible to human observers.

\subsubsection{Performance}
The efficacy of adversarial attacks is evaluated using different metrics, depending on the type of model being targeted. These metrics quantify the success rate of perturbations in causing model errors.
\paragraph{Classification Models.}
For classification models, the Attack Success Rate (ASR) serves as the primary metric, defined as the percentage of adversarial examples that successfully cause misclassification. For targeted attacks, ASR measures the percentage of samples that are misclassified as the attacker's chosen target class. For untargeted attacks, it measures the percentage of samples that are simply misclassified (regardless of which incorrect class is chosen). Mathematically, ASR can be expressed as:
\begin{equation}
    \text{ASR} = \frac{\text{Number of successful adversarial examples}}{\text{Total number of adversarial examples}}\times 100\%.
\end{equation}
Higher ASR values indicate more effective attacks. When evaluating defense mechanisms, a lower ASR demonstrates greater robustness against adversarial perturbations.
\paragraph{Object Detection Models.}
For object detection models, attack performance is evaluated using metrics that address both localization and classification aspects. Common metrics include:
\begin{itemize}
    \item \textbf{Mean Average Precision (mAP) drop}: the reduction in mAP between clean and adversarial inputs.
    \item \textbf{Detection rate reduction}: The percentage decrease in correctly detected objects.
    \item \textbf{False positive increase}: The percentage increase in falsely detected objects.
    \item \textbf{Disappearance rate}: The percentage of objects that are completely missed by the detector after perturbation.
\end{itemize}
These metrics capture different aspects of disruption to object detection systems, with successful attacks typically causing significant degradation across multiple metrics.
\paragraph{Segmentation Models.}
For segmentation models, which predict pixel-wise class labels, attack performance is evaluated using segmentation-specific metrics:
\begin{itemize}
    \item \textbf{Intersection over Union (IoU) reduction}: The decrease in overlap between ground truth and predicted segmentation masks.
    \item \textbf{Mean IoU (mIoU) drop}: The average IoU reduction across all classes.
    \item Pixel accuracy decrease: The reduction in correctly classified pixels.
    \item \textbf{Boundary F1 score reduction}: The decrease in accuracy of object boundaries.
\end{itemize}
Effective attacks against segmentation models typically result in substantial decreases in these metrics, indicating significantly degraded segmentation quality.

\subsubsection{Transferability}
Transferability measures the effectiveness of adversarial examples when applied to models different from the one used to generate them. This property is particularly important for black-box attacks, where direct access to the target model is limited or unavailable. Transferability is typically evaluated by:
\begin{itemize}
    \item \textbf{Cross-model transferability}: Success rate when transferring between different model architectures (e.g., from ResNet~\cite{he2016deep} to VGG~\cite{simonyan2014very}).
    \item \textbf{Cross-task transferability}: Success rate when transferring between different tasks (e.g., from classification to object detection).
    \item \textbf{Cross-dataset transferability}: ASR when transferring between models trained on different datasets.
\end{itemize}
Highly transferable attacks are generally considered more powerful, as they indicate the exploitation of fundamental vulnerabilities rather than model-specific weaknesses. Transferability is quantified by calculating the ASR or performance degradation metrics on the transfer model compared to the source model.

\subsubsection{Imperceptibility}
Imperceptibility measures how difficult it is for humans to detect the adversarial perturbations, which is crucial for practical attacks that must remain unnoticed. Various Image Quality Assessments (IQAs) are used to evaluate imperceptibility:
\begin{itemize}
    \item \textbf{Peak Signal-to-Noise Ratio (PSNR)}: Measures the ratio between the maximum possible power of a signal and the power of corrupting noise. Higher PSNR values indicate greater similarity between the original and perturbed images.
    \item \textbf{Structural Similarity Index Measure (SSIM)}: Evaluates the similarity of two images based on luminance, contrast, and structure, with values ranging from -1 to 1~\cite{wang2004image}. SSIM values closer to 1 indicate greater similarity.
    \item \textbf{Fr\'echet Inception Distance (FID)}: Measures the distance between the feature representations of real and generated images~\cite{heusel2017gans}. Lower FID scores indicate greater perceptual similarity.
    \item \textbf{Learned Perceptual Image Patch Similarity (LPIPS)}: Utilizes deep neural networks to measure perceptual similarity, better aligning with human visual perception than traditional metrics~\cite{zhang2018unreasonable}. Lower LPIPS values indicate greater perceptual similarity.
\end{itemize}
Additionally, human perception studies are sometimes conducted to evaluate imperceptibility directly, as computational metrics do not always perfectly align with human visual perception. In these studies, participants are asked to distinguish between original and adversarial images, with higher success rates indicating less imperceptible perturbations.
\section{Pixel-Space Attacks}

Pixel-space attacks (PSAs) involve modifying pixel values within certain constraints to generate adversarial samples that can deceive machine learning models. These attacks represent the earliest category of adversarial methods proposed in the literature and continue to receive the most extensive attention from researchers. As this field has evolved, significant advancements have been made in improving both the transferability of these attacks across different models and their imperceptibility to human observers. This section provides a comprehensive review of the developments in PSA, mainly focusing on gradient-based attacks, examining their methodologies and the progressive improvements in their effectiveness and sophistication.

\subsection{Early Foundations and Momentum-Based Enhancements}

\begin{figure}[t]
    \centering
    \includegraphics[width=\linewidth]{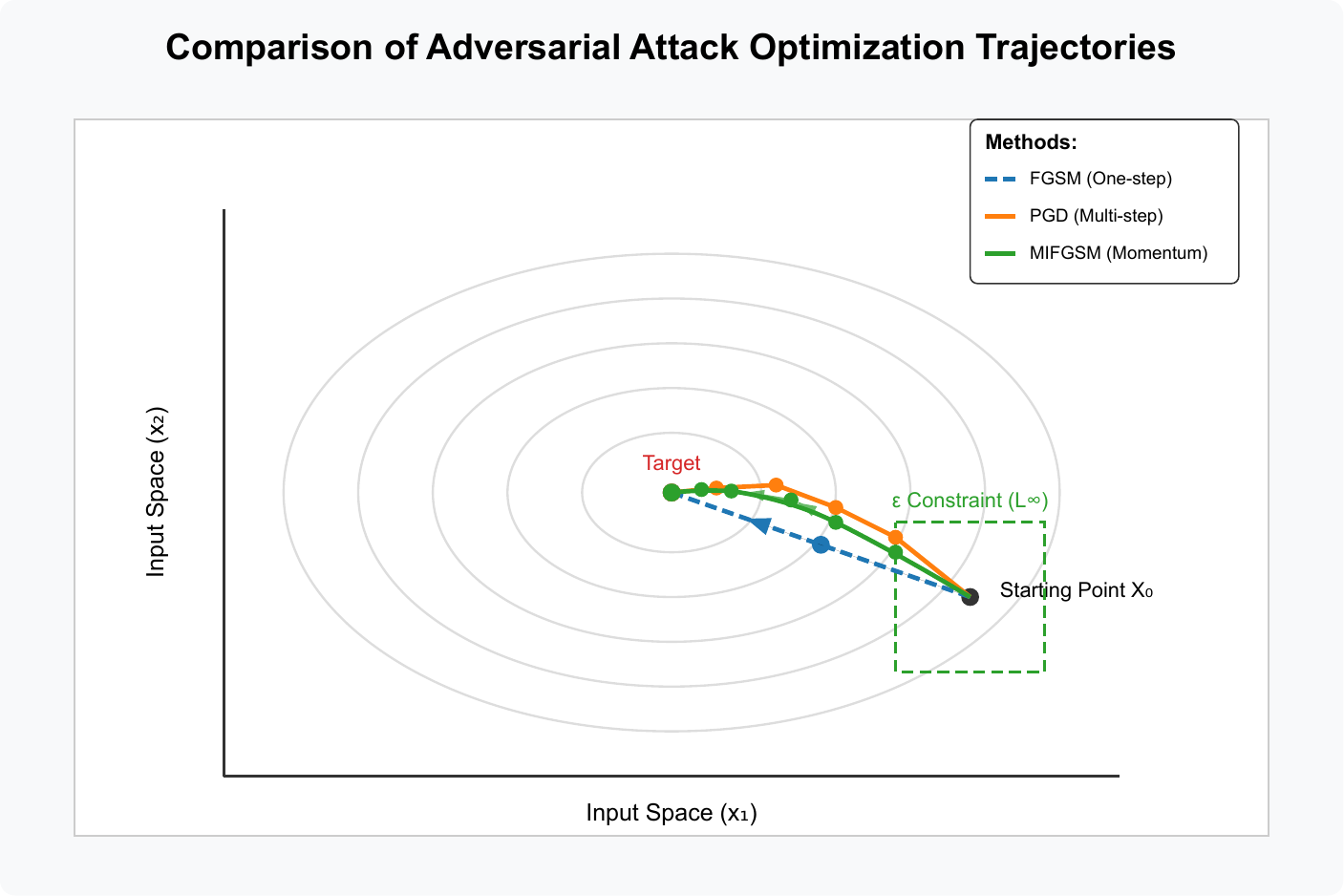}
    
\vspace{0.5em}
\resizebox{\textwidth}{!}{%
\begin{tabular}{c|c|c|c}
\hline
Method  & Iterations                  & Key Features                                                                                               & Update Formula \\ \hline
FGSM    & Single-step                 & \begin{tabular}[c]{@{}c@{}}Fast but limited effectiveness\\ Direct gradient approach\end{tabular}          & $x^{adv} = x + \epsilon\cdot\operatorname{sign}(\nabla_xJ(x,y))$            \\ \hline
PGD     & \multirow{2}{*}{Multi-step} & \begin{tabular}[c]{@{}c@{}}Stronger attack with projection\\ Can follow complex loss surfaces\end{tabular} & $x_{t+1}^{adv} = \Pi(x_t^{adv} + \alpha\cdot\operatorname{sign}(\nabla_xJ(x_t^{adv},y)))$            \\ \cline{1-1} \cline{3-4} 
MI-FGSM &                             & \begin{tabular}[c]{@{}c@{}}Better transferability\\ Avoids local noise \& oscillation\end{tabular}         & \begin{tabular}[c]{@{}c@{}}$g_{t+1} = \mu\cdot g_t + \nabla_{x}J(x_t^{adv},y)\|\nabla_xJ(x^{adv}_t,y)\|_1$\\ $x_{t+1}^{adv} = x_t^{adv} + \alpha\cdot\operatorname{sign}(g_{t+1})$\end{tabular}            \\ \hline
\end{tabular}%
}
    \caption[Visual optimization comparison of FGSM, PGD, and MI-FGSM.]{Visual optimization comparison of FGSM, PGD, and MI-FGSM.}
    \label{fig:chap2:vis-compare}
\end{figure}

As shown in Figure~\ref{fig:chap2:vis-compare}, the Fast Gradient Sign Method (FGSM)~\cite{goodfellow2014explaining} and its iterative variants (I-FGSM~\cite{kurakin2018adversarial}, PGD~\cite{madry2018towards}) represent the first generation of PSA. These methods employ the gradient of the given loss function with respect to the input image to determine the update of perturbation. Specifically, FGSM generates adversarial examples $x^{adv}$ from clean images $x$ by one step:
\begin{equation}
    x^{adv} = x + \epsilon \cdot \text{sign}(\nabla_{x} \mathcal{J}(x, y)),
\end{equation}
where $\nabla_{x} \mathcal{J}(x, y)$ is the gradient of the loss function $\mathcal{J}$ with respect to input $x$ for true label $y$, and $\epsilon$ is the perturbation magnitude.

A significant progress was made by the Momentum Iterative Fast Gradient Sign Method (MI-FGSM)~\cite{dong2018boosting}. This approach incorporated momentum terms into the iterative attack optimization process, substantially improving transferability across different victim models. The update process of MI-FGSM is:
\begin{equation}
    \begin{aligned}
        g_{t+1} = \mu \cdot g_t + \frac{\nabla_x J(x_t^{adv}, y)}{||\nabla_x J(x_t^{adv}, y)||_1},\quad
        x_{t+1}^{adv} = x_t^{adv} + \alpha \cdot \text{sign}(g_{t+1}),
    \end{aligned}
\end{equation}
where $\mu$ is the momentum decay factor, $g_t$ is the accumulated gradient at iteration $t$, and $\alpha$ is the step size, or generally speaking, learning rate.

MI-FGSM's high performance laid the foundation for further momentum-based enhancements. Nesterov momentum~\cite{lin2020nesterov} was considered to improve MI-FGSM (N-MI-FGSM), which employs a ``look ahead'' gradient to more efficiently move away from suboptimal regions. Their method computes gradients at an estimated future position instead of the current position:
\begin{equation}
    \tilde{x}_t^{adv} = x_t^{adv} + \alpha \cdot \mu \cdot g_t, \quad
    g_{t+1} = \mu \cdot g_t + \frac{\nabla_x J(\tilde{x}_t^{adv}, y)}{||\nabla_x J(\tilde{x}_t^{adv}, y)||_1}.
\end{equation}

\subsection{Critique of the Sign Function and Gradient Refinement}

The aforementioned methods have established the fundamental framework for adversarial attacks. However, these approaches have been questioned by several researchers, particularly regarding the use of the sign function. Critics argue that this function discards the amplitude information present in the gradient, which may not consistently yield optimal adversarial examples.

Research into gradient function limitations appears in~\cite{cheng2021fast}, where Taylor expansion analysis revealed directional inefficiencies of the sign operation, leading to their proposed Fast Gradient Non-sign Method (FGNM) with corresponding mathematical improvements. Complementary work by~\cite{yuan2024adaptive} established that direct gradient scaling techniques surpass sign-based approaches in transferability metrics while maintaining visual imperceptibility standards.

Researchers in~\cite{cheng2021fast} examined previous approaches, uncovering sign function inefficiencies and subsequently introducing FGNM. Their Taylor expansion analysis revealed limitations in sign-based methods while proposing mathematical improvements for attack optimization.
Following this work, scholars developed direct gradient scaling techniques to replace sign functions~\cite{yuan2024adaptive}. Experiments verified these methods enhanced cross-model transferability of adversarial examples while maintaining necessary imperceptibility characteristics.

Raw Gradient Descent (RGD)~\cite{yang2023rethinking} was proposed, which entirely removes the sign operation in attack framework. By reformulating the optimization problem from constrained to unconstrained, RGD utilizes the raw gradient directly:
\begin{equation}
    x_{t+1}^{adv} = \Pi_{B_\epsilon(x)} \left( x_t^{adv} + \alpha \cdot \frac{\nabla_x J(x_t^{adv}, y)}{||\nabla_x J(x_t^{adv}, y)||_\infty} \right),
\end{equation}
where $\Pi_{B_\epsilon(x)}$ is the projection operation onto the $\epsilon$-ball centered at $x$. Their comprehensive experiments showed that RGD consistently outperforms PGD across various settings.

\subsection{Advanced Optimization and Adaptive Dynamics}

The integration of advanced optimization techniques into the well-established attack frameworks represents another advancement. Nadam optimizer~\cite{zhang2023boosting} was incorporated into MI-FGSM, combining adaptive step size with look-ahead momentum to enhance effectiveness and transferability. The update rule is modified as:
\begin{equation}
    \hat{m}_t = \frac{m_t}{1-\beta_1^t},\quad
    \hat{v}_t = \frac{v_t}{1-\beta_2^t},\quad
    x_{t+1}^{adv} = x_t^{adv} + \alpha \cdot \text{sign}\left(\frac{\beta_1 \cdot \hat{m}_t + (1-\beta_1) \cdot g_t}{\sqrt{\hat{v}_t} + \epsilon}\right),
\end{equation}
where $m_t$ and $v_t$ are the first and second moment estimates of the gradient, and $\beta_1$ and $\beta_2$ are the decay rates.

Researchers incorporated the AdaBelief optimizer into the attack framework, termed AB-FGSM~\cite{wang2021generalizing}. This method dynamically adjusts $\alpha$ according to the estimated confidence in gradient calculations, resulting in notable cross-model transferability against both standard and adversarially-trained defensive systems.
In a parallel development, the Adam Iterative Fast Gradient Method (AI-FGM)~\cite{zhou2024study} harnesses Adam's momentum-based optimization and adaptive learning rates to construct more effective adversarial perturbations, demonstrating enhanced efficiency in penetrating model defenses.

To dismiss the adjustable step size, a general framework was presented~\cite{tao2023adapting}, which can be implemented within aforementioned methods, such as MI-FGSM, N-MI-FGSM, etc. The proposed architecture can ensure better convergence and stability.
The concept was extended through the introduction of a non-monotonic adaptive momentum coefficient combined with variable step-size methodology~\cite{long2024convergence}, accompanied by formal theoretical guarantees concerning regret limitations for convex functional spaces.

An innovative approach was introduced, employing variable step sizes that evolves throughout the attack process~\cite{huang2024using}. Their method specifically targets the common issue in iterative attacks such as I-FGSM, where transferability effectiveness tends to diminish with increasing iterations. By strategically adjusting and optimizing the utilization of gradient steps, their technique successfully preserves high levels of cross-model transferability across the entire optimization sequence, offering a solution to a significant limitation in traditional iterative attack methodologies.

``WITCHcraf'' was proposed~\cite{chiang2020witchcraft}, a novel technique that incorporates randomized step sizes into the PGD framework. This approach effectively reduces initialization sensitivity while boosting performance efficiency without imposing additional computational burden. The research demonstrated that strategically randomizing the step sizes in the attack algorithm yields substantial improvements in successful attack rates. Despite its simplicity, this modification to traditional PGD proved to be remarkably effective in enhancing adversarial attack capabilities.

\subsection{Transferability Enhancements and Spatial Considerations}

Improving the transferability of adversarial examples across different models has been a central focus of recent research. Beyond the momentum-based methods discussed earlier, several innovative approaches have emerged.

The research work presented in~\cite{wang2022enhancing} proposed a novel approach called Spatial Momentum Iterative FGSM (SMI-FGSM), which extends traditional adversarial attack methods by incorporating spatial-wise gradient accumulation alongside temporal momentum. This innovative technique takes into account contextual gradient information within images, resulting in more stable gradient update processes across various model architectures and datasets. By leveraging both spatial and temporal dimensions for momentum accumulation, SMI-FGSM demonstrates enhanced transferability capabilities compared to previous methods that relied solely on temporal momentum strategies.

The technique known as Scheduled Step Size and Dual Example (SD) was introduced in~\cite{zhang2023improving}, which employs dynamic step size adjustment alongside dual examples to concentrate perturbations in proximity to benign samples. This methodology significantly enhances cross-model transferability by effectively preventing optimization processes from deviating excessively from the original sample distribution. Through this strategic approach, the researchers demonstrated improved attack efficiency while maintaining perturbation relevance.

The work presented in~\cite{fang2023adversarial} introduces HE-MI-FGSM, a novel attack leveraging histogram equalization techniques to mitigate overfitting issues and boost black-box transferability. Through effective perturbation distribution regularization, this approach maintains its attacking capabilities while significantly improving generalization to unknown defensive models.

Research presented in~\cite{jang2022strengthening} introduced a novel integration of Lookahead FGSM with Self-CutMix techniques to tackle the transferability limitations commonly observed under adversarial training scenarios. This approach enhances attack performance by intelligently utilizing internal patches from the target image, thereby preserving critical visual information and substantially improving success rates in black-box attack environments.

A different methodology described in~\cite{zhu2023ICCV} introduced the Gradient Relevance Attack framework, which implements adaptive direction correction mechanisms during the iterative perturbation generation process. This technique effectively minimizes gradient fluctuations by incorporating neighborhood information to adjust gradient relevance, resulting in remarkably high success rates against sophisticated defensive systems in black-box scenarios.

The work in~\cite{wan2023average} established an innovative Average Gradient-Based Adversarial Attack methodology that constructs and maintains a dynamic collection of adversarial examples throughout the attack process. By averaging gradient information across multiple iterations, this approach successfully mitigates the noise and instability issues inherent in traditional gradient updates, consequently enhancing the transferability of generated adversarial examples.

\subsection{Feature-Region and Specialized Techniques}

Numerous investigations have concentrated on specialized approaches for improving attack efficiency or minimizing detectability. The F-MIFGSM technique was developed~\cite{liu2020f}, which confines perturbations to specific feature areas through the application of convolutional and deconvolutional neural network layers. By focusing on the most important regions within images, this method enhances attack concealment while simultaneously preserving high success metrics.

A different approach called Fast Gradient Scaled Method (FGScaledM) was developed~\cite{xu2022fast}, which implements gradient scaling to reduce perceptibility without compromising attack effectiveness. This methodology demonstrated that precisely controlled gradient scaling mechanisms could generate adversarial examples with substantially reduced visual detectability.

Subsequently, researchers established the Scale-Invariant PGD (SI-PGD) methodology~\cite{xu2022scale}, employing angular characteristics rather than logits to maintain consistent attack performance despite logit rescaling operations. This particular technique proved especially powerful against defensive mechanisms that rely on input transformation strategies or model scaling techniques.

\subsection{Theoretical Advances and Comprehensive Frameworks}

In the field of adversarial attacks, substantial progress has been made toward establishing theoretical underpinnings and consolidated frameworks for gradient-oriented attack methodologies. Researchers have proposed interpretative models that explain how diverse attack techniques can be conceptualized as variations of gradient descent with unique adaptation protocols~\cite{tao2023adapting}. Such frameworks not only deliver convergence assurances but also elucidate the interconnections between seemingly disparate attack approaches.

Further theoretical advancements have been achieved in understanding the mathematical properties of adaptive momentum techniques in adversarial contexts. A rigorous analysis establishing theoretical boundaries for regret in these methods has provided formal validation for the efficacy of using variable momentum parameters during optimization processes in adversarial settings~\cite{long2024convergence}.

The research community has additionally produced evaluative research comparing various attack implementations. A methodical evaluation of Fast Gradient Sign Method variants has been conducted, examining their operational mechanisms, inherent limitations, and performance impacts on ImageNet-trained ResNet-50 architectures~\cite{lad2024fast}. These comparative investigations enhance our comprehension of the compromises inherent in different attack strategies and inform the development of more robust defensive measures.

\subsection{Summary}

Overall, PSAs have evolved from simple gradient-based methods to sophisticated optimization techniques that leverage advanced momentum, adaptive step sizes, and specialized refinements. The critique of the sign function has led to more effective gradient utilization, while transferability enhancements have improved attack success in black-box scenarios. These developments collectively represent a significant advancement in the field of adversarial machine learning, challenging the robustness of deep learning models in real-world applications.

\section{Physically Realizable Attacks}

\begin{figure}[t]
    \centering
    \includegraphics[width=\linewidth]{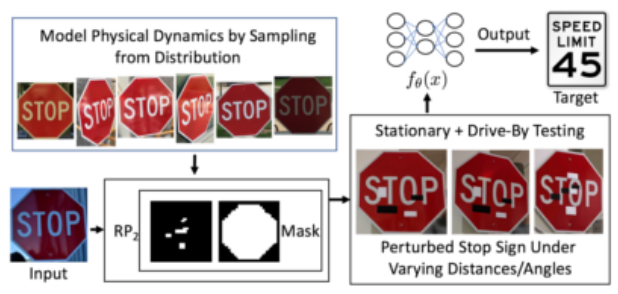}
    \caption{An example of physically realizable attacks~\cite{eykholt2018robust}, where the adversarial patch can easily fool the machine learning model.}
    \label{fig:chap2:adv-patch}
\end{figure}

Physically realizable adversarial attacks constitute a notable progression in the field of adversarial machine learning, transforming theoretical security vulnerabilities into concrete real-world threats. As shown in Figure~\ref{fig:chap2:adv-patch} in contrast to digital-domain perturbations, attacks implemented in physical environments must sustain their effectiveness across various real-world conditions including changes in lighting, shifts in viewpoint, and distortions from printing processes. This section examines the progression and refinement of these attack methodologies and their deployment across various computer vision applications.

\subsection{Early Foundations}

The emergence of physically implementable adversarial attacks was pioneered through the development of the Robust Physical Perturbations (RP$^2$) algorithm~\cite{eykholt2018robust}. This seminal research illustrated how strategically designed adhesive patterns could induce misclassification of traffic signage when observed across varying perspectives, distances, and illumination scenarios. The RP$^2$ methodology incorporated an optimization framework that considered real-world physical variations, imperfections in printing processes, and optical distortions, resulting in attack efficacy exceeding 80\% during practical field evaluations. This groundbreaking investigation confirmed the feasibility of adversarial manipulations extending beyond computational environments and emphasized potential vulnerabilities in critical safety systems such as self-driving vehicles.

In parallel developments, researchers introduced techniques for generating physically robust adversarial three-dimensional objects~\cite{athalye2018synthesizing}. This approach utilized the Expectation Over Transformation (EOT) framework to create adversarial examples maintaining effectiveness despite various transformations including rotational changes, spatial repositioning, and illumination differences. By incorporating these transformation models during the optimization procedure, the research produced 3D-printed artifacts that consistently deceived classification systems regardless of observation angle. A notable example featured a turtle model consistently misidentified as a rifle from virtually any perspective, demonstrating how digital vulnerabilities successfully transfer to tangible physical objects.

\subsection{Adversarial Patches and Stickers}
Following these initial efforts, the domain of adversarial patches witnessed substantial advancement. Research presented in~\cite{wei2022adversarial} introduced the concept of Adversarial Sticker attacks, which generated physically printable elements optimized for real-world applications. This methodology specifically tackled printing-related distortions and practical physical constraints, proving its efficacy across diverse applications including facial identification systems, traffic signage recognition, and image-based retrieval mechanisms. A particularly valuable contribution of this approach was its ability to maintain effectiveness in black-box scenarios, enabling successful attacks against systems without requiring access to their architectural specifications.

The field progressed significantly when researchers conducted comprehensive evaluations of the RP$^2$ approach across various environmental conditions~\cite{eykholt2018robust}. Their investigations revealed how tactically positioned monochromatic adhesives could reliably cause autonomous systems to misinterpret stop signs as speed limit indicators, exposing critical vulnerabilities in vehicular automation technologies. This investigation highlighted the significance of environmental resilience, demonstrating that physical adversarial manipulations could sustain their effectiveness across varying observation distances, perspectives, and illumination settings.

A notable progression in adversarial methodologies emerged with the development of Universal Camouflage Patterns (UCP)~\cite{huang2020universal}. Diverging from previous techniques that focused on specific targets, UCP generated adversarial patterns that remained effective against object detection systems across multiple contexts. This investigation introduced AttackScenes, a three-dimensional simulation environment that replicated real-world settings to validate attack performance, effectively connecting digital optimization processes with physical implementation challenges. The developed patterns achieved approximately 75\% degradation in detection capabilities when applied to vehicles in authentic testing environments.

The quest for developing inconspicuous yet functional adversarial elements was addressed through the introduction of Naturalistic Physical Adversarial Patches (NPAP)~\cite{HU2021ICCV}. This novel approach utilized generative adversarial networks to create visually convincing patches that appeared ordinary to human observers while maintaining their adversarial impact against object detection frameworks. By employing pre-established GANs such as BigGAN and StyleGAN, NPAP generated patches mimicking natural textures and designs, substantially enhancing the covertness of physical attacks while sustaining attack efficacy rates exceeding 80\% against YOLOv3 detection systems.

In specialized application domains, research presented in~\cite{cheng2022physical} introduced adversarial patches specifically targeting monocular depth estimation technologies. These specially optimized elements were designed to disrupt depth perception in autonomous driving systems, inducing measurement errors of up to 6 meters with success rates of 93\%. This work expanded the vulnerability landscape beyond traditional classification and detection tasks, demonstrating significant weaknesses in three-dimensional perception systems that are essential for navigation and obstacle detection functions.

\subsection{3D Textures and Wearable Attacks}
The transition from basic 2D adversarial patches to sophisticated 3D textured models marks a crucial advancement in physically realizable attack methodologies. Research introduced the Adversarial Textured 3D Meshes (AT3D) framework~\cite{yang2023towards}, which specifically targeted facial recognition technologies. This innovative approach tackled the intricate problem of designing adversarial textures that retain their effectiveness when applied to three-dimensional objects with diverse geometric properties. Through optimization of texture patterns that preserved their adversarial characteristics when mapped onto 3D facial meshes, researchers demonstrated highly effective attacks against commercial facial recognition frameworks under authentic operating environments, achieving ASR above 90\% in real-world testing scenarios.

The development of wearable adversarial elements represents a particularly significant security concern due to their mobility and practical implementation potential. The Adversarial Clothing Textures (AdvCaT) methodology~\cite{hu2023CVPR} addressed the specific challenges presented by non-rigid surfaces such as garments. This technique utilized advanced 3D modeling combined with Voronoi parameterization to generate robust textures capable of maintaining their adversarial properties despite the deformations experienced by clothing during normal body movements. To minimize the gap between digital simulations and physical implementation, this research incorporated fabrication constraints and color calibration protocols, ultimately achieving 90\% attack efficacy against person detection systems across diverse environmental conditions.

The introduction of Legitimate Adversarial Patches (LAPs)~\cite{tan2021legitimate} brought forward the critical concept of perceptual reasonableness alongside attack effectiveness. The research employed a dual-phase training methodology to create patches that appeared contextually natural while simultaneously retaining their capability to undermine object detection systems. By addressing the human perception aspect, this work overcame a significant limitation of previous adversarial patch designs that were visually conspicuous and readily identifiable as malicious elements. The LAPs approach demonstrated over 70\% ASR while substantially reducing human suspicion metrics compared to conventional adversarial patch implementations.

Comprehensive evaluation research~\cite{zarei2022adversarial} provided thorough assessment of three-dimensional physical adversarial attacks, contrasting the performance of traditional 2D patches against complete 3D adversarial objects. This investigation examined critical variables including illumination variations, positional alterations, and material characteristics, establishing that 3D adversarial objects maintained superior ASR (85\% compared to 67\%) under varied observational conditions relative to their 2D counterparts. This systematic analysis established valuable benchmarks for assessing physical attack robustness and emphasized the enhanced adaptability of three-dimensional adversarial formations to environmental variations.

\subsection{Dynamic and Optical Attacks}
The field of adversarial attacks witnessed a significant advancement with the emergence of dynamic projection-based techniques. The concept of Short-lived Adversarial Perturbations (SLAP)~\cite{lovisotto2021slap} introduced a methodology utilizing projected light patterns to create temporary adversarial elements on objects. This innovative approach provided exceptional adaptability, enabling real-time adjustment to environmental fluctuations and moving targets. These projected adversarial elements achieved an 87\% success rate in causing misclassification of traffic signage while leaving no enduring physical evidence, thus presenting substantial challenges for detection and countermeasures. The effectiveness of SLAP was demonstrated against both classification algorithms and object detection frameworks, with pattern optimization tailored for specific target models.

The optical attack landscape was further developed through EvilEye~\cite{han2023don}, an approach that implemented transparent display technology to generate dynamic optical disturbances. This research formalized the physical manifestation of digital attacks, establishing a versatile framework for real-time adversarial pattern deployment. Particularly noteworthy was EvilEye's performance in safety-critical applications, where it achieved ASR exceeding 90\% against perception systems in surveillance and autonomous driving contexts. The method's capability to adjust perturbation intensity based on ambient conditions ensured consistent effectiveness across diverse lighting scenarios and distances, highlighting its remarkable environmental adaptability.

Another innovative direction in this domain emerged with Reflected Light Adversarial Attack (RFLA)~\cite{wang2023ICCV}, which employed strategically designed light reflection patterns to deceive computer vision systems. By optimizing reflection characteristics to maintain robustness under varied environmental conditions, RFLA demonstrated an impressive 99\% ASR against image classification frameworks without requiring physical alterations to target objects. This methodology enabled nearly imperceptible attacks that presented exceptional challenges for detection and defense mechanisms, as the adversarial elements existed solely as ephemeral light patterns rather than permanent physical modifications.

The expansion into infrared-domain attacks broadened vulnerability exploitation to include thermal imaging technologies. Research introduced physically adversarial infrared patches~\cite{wei2023CVPR} specifically designed to target thermal cameras deployed in surveillance systems and autonomous vehicles. These patches modified thermal distributions to generate adversarial patterns within the infrared spectrum, achieving over 80\% success rates against detectors processing thermal imagery for pedestrian and vehicle identification. Through optimization of both patch configuration and positioning, this approach maintained effectiveness across temperature variations and viewing conditions, revealing vulnerabilities in systems engineered for operation in low-light environments and adverse weather scenarios.

\subsection{Stealth Optimization Techniques}
With the advancement of detection and defense strategies, the research community began to place greater emphasis on developing stealthier physical attack methodologies. The Dual Attention Suppression (DAS) attack was proposed~\cite{wang2021dual}, which employed a simultaneous optimization approach to circumvent both machine learning model attention mechanisms and human visual perception. This innovative technique generated contextually appropriate camouflage patterns that maintained adversarial effectiveness while appearing more natural to human observers, resulting in substantially improved imperceptibility metrics. Experimental validation demonstrated the DAS attack's efficacy across classification and object detection systems, with real-world physical implementations confirming its resilience under authentic operational conditions.

Further refinements in adversarial patch inconspicuousness emerged through sensitivity mapping techniques. A novel approach utilizing sensitivity maps was developed~\cite{wang2024stealthy} to identify and exploit the most susceptible regions within object detection architectures while simultaneously reducing patch dimensions and visual prominence. By strategically concentrating adversarial perturbations at high-sensitivity model locations, this methodology achieved patch size reductions of approximately 60\% while sustaining ASR exceeding 75\%. This significant dimensional decrease substantially enhanced the stealthiness factor by rendering the adversarial elements considerably less detectable to human observers without sacrificing their effectiveness against computational systems.

Complementary research explored sparse adversarial patterning strategies. The Maximum Aggregated Region Sparseness (MARS) approach~\cite{zhao2025local} was formulated to minimize and strategically localize attack regions on three-dimensional objects. Through the calculated placement of compact adversarial patterns at optimal positions, MARS achieved attack performance comparable to full-surface perturbations while substantially decreasing the modified object area. This methodology proved particularly advantageous for 3D object attacks, where precise positioning of small, unobtrusive adversarial elements at key viewpoints established an optimal balance between concealment and effectiveness.

Generative adversarial network (GAN) based natural-looking patches constituted another significant advancement in stealth optimization. As previously examined, the NPAP methodology~\cite{HU2021ICCV} employed generative modeling techniques to create adversarial patches resembling authentic environmental textures. Subsequent investigations further refined this approach, developing implementations capable of remarkably convincing visual similarity to natural elements such as geological formations, plant matter, or textile patterns while preserving their adversarial characteristics. These naturalistic techniques substantially reduced human detection probability while maintaining high success rates against computer vision systems.

\nocite{qian2024semi,li2024threats,qian2025perspective,zhao2024survey,hu2025syntactic,li2025achieving}

The introduction of distillation-enhanced optimization techniques for physical adversarial patches~\cite{liu2025distillation} represented an additional advancement in this domain. This methodology leveraged knowledge distillation principles to transfer adversarial properties between models, yielding patches with approximately 20\% improved attack effectiveness alongside enhanced environmental integration capabilities. By optimizing patches to simultaneously emulate surrounding textures while maintaining their adversarial functionality, this approach achieved improved equilibrium between visual imperceptibility and attack efficacy, particularly for deployment scenarios involving visually complex environments.

\subsection{Specialized Applications and Future Directions}
When extended to specialized domains, physically realizable attacks have uncovered unique vulnerabilities specific to various contexts. Research in aerial detection systems has evaluated the effectiveness of adversarial patches against drone and satellite imaging technologies~\cite{lian2022benchmarking}. This investigation addressed the distinctive challenges presented by aerial viewpoints, including unusual observation angles and significant distances. The developed attack methods demonstrated resilience against variations in scale and environmental conditions inherent to aerial surveillance, illustrating how physical attacks can be adapted to specialized operational contexts.

Research has also introduced innovative attack vectors such as Out-of-Bounding-Box Triggers~\cite{lin2024out}, which represent a covert approach that positions adversarial elements outside the conventional bounding box regions utilized by object detection algorithms. Through the implementation of feature guidance techniques and unified adversarial patch gradient descent methodology, researchers developed inconspicuous triggers capable of inducing detection failures while maintaining visual separation from targeted objects. This strategy achieved ASR exceeding 85\% in real-world environments while substantially reducing human attention metrics compared to conventional adversarial patch implementations.

An emerging trend in this field involves the integration of multiple attack modalities. Contemporary approaches combine static physical patterns with dynamic light projections or incorporate adversarial textures with specialized materials affecting different sensing capabilities (including visual, infrared, and LiDAR systems). These multi-modal methodologies create significant obstacles for defensive mechanisms by simultaneously exploiting vulnerabilities across diverse sensing technologies and processing frameworks.

Current research increasingly emphasizes attack transferability across different computer vision tasks. While initial approaches targeted specific vision functions independently, recent investigations explore adversarial patterns that concurrently impact multiple vision operations. For instance, attacks initially designed for classification purposes have been successfully extended to compromise detection and segmentation functions without additional optimization, highlighting fundamental vulnerabilities shared across vision processing pipelines. This cross-task transferability substantially increases the potential threat of physical attacks in integrated vision systems.

Detailed literature reviews~\cite{wei2024physical,wei2022visually,guesmi2023physical} have methodically classified these diverse attack strategies, identifying patterns, limitations, and future research directions in physically realizable adversarial attacks. These comprehensive analyses emphasize the accelerating technical advancements in attack methodologies, the broadening range of affected vision applications, and increasingly sophisticated concealment techniques. They also underscore persistent challenges, including effectiveness-concealment trade-offs, disparities between digital optimization and physical implementation, and requirements for standardized evaluation frameworks to assess system robustness against physical attacks.

In conclusion, physically realizable adversarial attacks have progressed from fundamental proof-of-concept demonstrations to sophisticated, concealed, and resilient attack vectors targeting various computer vision applications. From initial algorithms to advanced generative approaches, three-dimensional textured objects, and dynamic optical perturbations, these methodologies consistently demonstrate the vulnerability of vision systems to adversarial manipulation in real-world environments. The ongoing advancement of these techniques, combined with their increasing imperceptibility and robustness, presents substantial challenges for security and reliability in vision-based systems deployed in critical applications.
\section{Latent-Space Attacks}

\begin{figure}
    \centering
    \includegraphics[width=\linewidth]{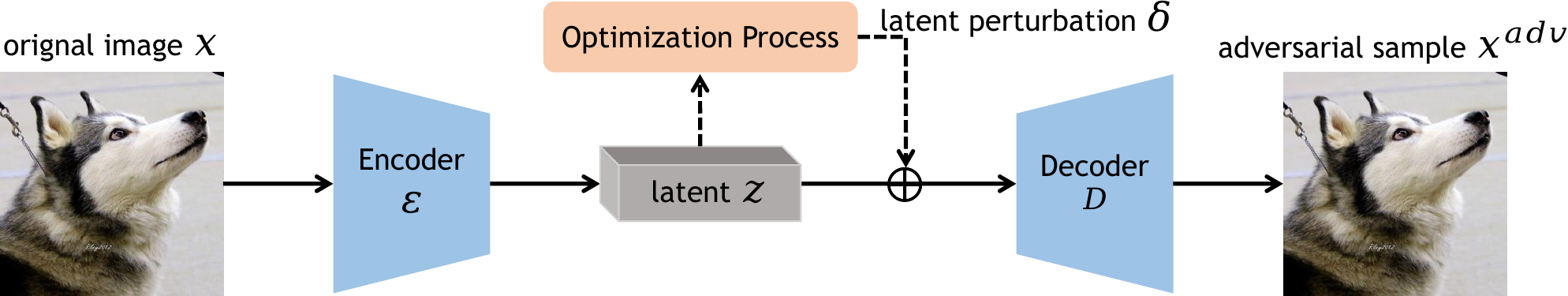}
    \caption{General paradigm of latent space attacks. The process flows from original samples through an encoder to latent space, where optimized perturbations are applied, followed by decoder transformation to generate deceptive yet visually similar adversarial examples. This attack methodology leverages the semantic structure of the latent space to achieve more efficient and targeted adversarial manipulations.}
    \label{fig:chap2:lsa}
\end{figure}

Feature representation-based adversarial approaches form a sophisticated category of attack methods that target the internal feature spaces of deep neural networks rather than directly modifying input elements. As summarized in Figure~\ref{fig:chap2:lsa}, these techniques exploit vulnerabilities in high-level abstractions within models, enabling the generation of adversarial samples with enhanced cross-model transferability, greater semantic significance, and improved resilience against conventional defensive techniques.

\subsection{Early Foundations}

The investigation into LSAs gained momentum during 2017 -- 2019, representing a paradigm shift from conventional pixel-space modifications toward manipulations within neural networks' internal representations. The groundbreaking work on LatentPoison~\cite{creswell2017latentpoison} demonstrated the feasibility of generating adversarial examples through perturbations in deep variational autoencoders' latent representations. This approach implemented additive modifications directly to latent encodings, resulting in incorrect model predictions while preserving visual fidelity to the original inputs. This research established that exploiting latent vulnerabilities could facilitate more covert attacks with reduced perceptible artifacts.

Subsequent research expanded this direction when researchers proposed AdvGAN++~\cite{jandial2019advgan}, which utilized generative adversarial networks to create adversarial examples through latent feature manipulation rather than direct image modification. This approach differentiated itself from earlier GAN-based techniques operating in pixel space by specifically targeting internal feature representations, thereby achieving superior attack effectiveness while maintaining perceptual similarity. This development illustrated the potential of generative frameworks to systematically explore and exploit vulnerabilities within latent spaces.

Concurrent developments included the introduction of Feature Space Perturbations~\cite{inkawhich2019feature}, a methodology focused on improving adversarial transferability through strategic alignment of feature representations with target class characteristics. This technique identified the layers most susceptible to feature-level manipulations and demonstrated that latent-space perturbations could produce adversarial examples with enhanced cross-model generalization compared to conventional input-space approaches.

In response to these emerging threats, defensive strategies began to materialize, exemplified by the Latent Adversarial Defence (LAD)~\cite{zhou2019latent} framework, which targeted latent space vulnerabilities through decision boundary-oriented generation techniques. LAD operated by creating adversarial examples through feature perturbations along decision boundaries, subsequently reconstructing inputs for adversarial training purposes. This early defensive approach highlighted the growing recognition that securing latent spaces required specialized countermeasures beyond traditional input-space protection mechanisms.

\subsection{Feature Distribution and Manipulation Approaches}

The timeframe 2019 -- 2021 witnessed remarkable advancements in LSA strategies, particularly those focusing on manipulating feature distributions and semantically significant representations. The Feature Distribution Attack (FDA)~\cite{inkawhich2020transferable} emerged as a novel approach that deliberately altered class-specific feature distributions across multiple network layers, resulting in highly transferable adversarial examples. This methodology specifically targeted statistical characteristics within latent features, forcing them to deviate from their original class distribution toward designated target classes, thus demonstrating neural networks' susceptibility to distribution-based manipulations.

A significant breakthrough occurred with the introduction of LAFEAT~\cite{yu2021lafeat}, which pioneered gradient-oriented attacks specifically designed for latent feature manipulation. This technique utilized latent feature gradients to construct adversarial examples capable of circumventing even robust defense mechanisms, including models fortified through adversarial training. Research findings revealed that networks considered ``robust'' in input space remained vulnerable through their internal representations, highlighting a critical weakness in existing defensive approaches. By directly targeting intermediate features, this methodology achieved superior success rates against protected models compared to conventional attack vectors.

Research into semantic manipulation within latent spaces gained significant traction during this period. A framework for generating semantic adversarial examples through feature manipulation~\cite{wang2023generating} utilized disentangled latent codes to create subtle yet interpretable adversarial examples. This approach identified and altered specific semantic attributes such as object shape, color, and textural properties within latent representations to induce misclassification while preserving visual coherence. Complementary research~\cite{dunn2020semantic} presented methods for semantic adversarial perturbations using learned representations, employing generative adversarial networks to manipulate latent activations affecting high-level visual characteristics.

The vulnerability of internal layers in adversarially trained models received comprehensive examination through the development of LSA and Latent Adversarial Training (LAT)~\cite{kumari2019harnessing}. This research demonstrated that models specifically trained to withstand input-space perturbations nevertheless remained susceptible to manipulations of their latent representations. LAT addressed this vulnerability by implementing adversarial training directly within the latent space, establishing one of the earliest comprehensive defense mechanisms against LSAs.

Computational efficiency concerns in latent-space adversarial training were addressed through the Single-step Latent Adversarial Training (SLAT)~\cite{park2021reliably} methodology. Unlike multi-step alternatives requiring substantial computational resources, SLAT manipulated latent gradients in a single step to enhance adversarial robustness efficiently, substantially reducing computational demands while maintaining defensive effectiveness. This innovation rendered latent-space defenses considerably more feasible for practical implementation in resource-constrained environments.

\subsection{Generative Models and Semantic Control}

The period from 2021 to 2023 witnessed significant advancements in LSAs, incorporating sophisticated generative models and enhanced semantic manipulation techniques. Research presented in~\cite{upadhyay2021generating} introduced a novel methodology for creating out-of-distribution adversarial examples through manipulation of latent space. This approach utilized $\beta$-VAEs to modify latent representations, producing adversarial samples that extended beyond conventional data distributions while preserving visual integrity. The work established how latent space modifications could produce semantically consistent examples that challenged traditional adversarial constraints.

Further developments emerged with the introduction of concept-based adversarial attacks as described in~\cite{schneider2022concept}. This methodology manipulated latent activations in deeper network layers to generate adversarial examples through semantic concept interpolation. By targeting high-level semantic features, this approach demonstrated effectiveness against both machine learning classifiers and human perception, illustrating how LSAs could simultaneously exploit vulnerabilities in both computational and human recognition systems.

A crucial breakthrough occurred with the application of diffusion models for latent space manipulations. The research in~\cite{wang2023semantic} established Semantic Adversarial Attacks using Diffusion Models, implementing Semantic Transformation (ST) and Latent Masking (LM) techniques for precise semantic control. The generative capabilities of diffusion models enabled fine-grained manipulation of semantic attributes while maintaining image fidelity. Compared to previous GAN-based methodologies, this diffusion-based framework demonstrated enhanced semantic richness and improved attack transferability.

The Latent Magic framework described in~\cite{zheng2023latent} investigated adversarial examples created within the semantic latent space of Stable Diffusion models. This research introduced innovative metrics for quantifying both attack effectiveness and cross-model transferability, demonstrating that perturbations in rich semantic latent spaces could generate adversarial examples with superior transferability compared to conventional pixel-space perturbations. Experimental evidence confirmed that latent representations from advanced generative models could be leveraged to create highly effective cross-architecture attacks.

Research on the GLASSE framework~\cite{clare2023generating} integrated GANs with genetic algorithms to systematically explore latent spaces for adversarial example generation. This evolutionary approach navigated through latent manifolds to identify regions producing effective adversarial samples, achieving 82\% success rates against external classification systems. The combination of generative modeling with evolutionary search techniques demonstrated how complex latent spaces could be methodically explored to locate vulnerable regions susceptible to adversarial manipulation.

Additional work in~\cite{wang2022semantic} enhanced semantic preservation in adversarial attacks by combining autoencoder architectures with genetic algorithms. This methodology extracted latent representations via autoencoders and subsequently applied genetic algorithms to modify these representations while maintaining semantic integrity. The approach preserved higher semantic fidelity compared to previous techniques while delivering competitive ASR, emphasizing the growing importance of ensuring adversarial perturbations remain semantically coherent with original inputs.

\subsection{Advanced Geometric Understanding and Unrestricted Attacks}

Recent advances in latent-space adversarial techniques have increasingly emphasized more nuanced geometric understanding of latent representations and unconstrained adversarial manifolds. Research by~\cite{shukla2023generating} introduced an innovative approach utilizing GAN-based modifications that thoroughly examines geometric characteristics of embedded features. This methodology strategically shifts feature representations toward incorrect class convex hulls, effectively circumventing input-space noise limitations while maintaining visual authenticity. Through comprehensive geometric evaluation, feature visualization techniques, and analysis of class activation patterns, the research illustrated how precisely targeted geometric manipulations within latent spaces yield highly successful adversarial outcomes.

The development of Direct Adversarial Latent Estimation (DALE)~\cite{dale2024direct} offered new perspectives on decision boundary complexity assessment in black-box computational models. This framework employs variational autoencoders to generate and navigate through latent representations, producing adversarial samples for robustness evaluation. While primarily developed as an assessment tool rather than an attack mechanism, this investigation provided crucial insights regarding how decision boundaries in latent space influence model vulnerabilities, establishing theoretical foundations for understanding the variable effectiveness of latent perturbation techniques.

A breakthrough in unrestricted adversarial methodologies emerged with the introduction of Manifold-Aided Adversarial Examples~\cite{li2024transcending}. This technique leverages supervised generative architectures to manipulate semantic characteristics within latent spaces through adversarial manifolds, generating examples that extend beyond conventional perturbation limitations. By effectively separating semantic from non-semantic attributes, the approach creates adversarial instances with legitimate natural semantics that remain difficult to identify through standard detection methods. This represents an evolution toward adversarial samples that maintain complete semantic integrity while achieving substantial attack effectiveness.

The introduction of Semantic-Consistent Attacks (SCA)~\cite{pan2024sca} brought forth an exceptionally efficient framework for unrestricted adversarial attacks preserving semantic integrity. SCA implements diffusion inversion techniques to transform images into latent representations, modifies semantics under multimodal language model guidance, and ensures minimal distortion effects. This balanced approach combines adversarial optimization with semantic realism, demonstrating how contemporary generative models can be utilized to produce adversarial examples that maintain natural appearances while successfully deceiving target systems.

Research presented in~\cite{cao2022adversarial} proposed an Adversarial Attack Algorithm utilizing Edge-Sketched Features from Latent Space (LSFAA), which successfully eliminates input-level iterative processes for more efficient adversarial generation. Through training specialized feature extraction mechanisms, LSFAA rapidly produces adversarial examples by manipulating edge-based latent characteristics, illustrating how LSA methodologies can be optimized for computational efficiency while sustaining high success rates.

\subsection{Theoretical Insights and Properties}

Moving beyond individual attack strategies, numerous research efforts have offered valuable theoretical perspectives on LSA characteristics. Research presented in~\cite{casper2022robust} illustrated how adversaries targeting robust features could function as tools for interpretability, exposing model weaknesses through manipulation of high-level features in latent space. This investigation revealed that targeted, universal, and black-box latent space attacks could effectively highlight fragile semantic connections learned by neural networks, thus establishing a significant link between vulnerability to adversarial examples and model interpretability.

\nocite{guo2024generative,qian2025t2icount,lei2024instant}

Several studies have thoroughly investigated the cross-model transferability of LSA. Research in~\cite{inkawhich2020transferable} and~\cite{inkawhich2019feature} established that perturbations applied to latent representations generally demonstrate superior transferability across different model architectures compared to input-domain attacks. This enhanced transfer capability arises because diverse neural networks tend to develop comparable high-level feature representations despite architectural variations, making LSA approaches particularly potent in black-box attack scenarios.

The framework of Latent Manifold Adversarial Examples (LMAEs) introduced in~\cite{qian2021improving} examined vulnerabilities in latent distributions from both local and global perspectives. By strategically modifying latent distributions and implementing manifold-aware adversarial training techniques, this approach significantly improved model resilience against diverse attack vectors. The theoretical analysis provided connections between latent-space vulnerabilities and the geometric characteristics of feature manifolds, offering explanations for why certain latent space regions exhibit greater susceptibility to adversarial manipulation.

Research documented in~\cite{casper2021one} showcased the generation of universal, physically-implementable adversarial features through latent representation manipulation. Utilizing deep generative models alongside specialized optimization procedures, this work crafted interpretable attacks exploiting feature-class associations within latent space, demonstrating how semantic-level modifications could translate into practical threats against computer vision systems.

The relationship between LSAs and semantic coherence has emerged as a consistent theme in contemporary research. The semantic adversarial attack methodology proposed in~\cite{joshi2019semantic} employed parametric transformations in latent space, demonstrating how generative model latent spaces could be utilized to alter semantic attributes such as environmental conditions or facial characteristics. This research emphasized how semantically meaningful latent space transformations could generate adversarial outcomes that appear natural to human observers while consistently deceiving machine learning classifiers.

\subsection{Future Directions}

LSAs have evolved from simple perturbations in VAE and GAN latent codes to sophisticated manipulations leveraging advanced generative models and semantic control mechanisms. The progression from early works like LatentPoison~\cite{creswell2017latentpoison} and AdvGAN++~\cite{jandial2019advgan} to recent approaches using diffusion models~\cite{wang2023semantic} and adversarial manifolds~\cite{li2024transcending} reflects a deepening understanding of how internal feature representations can be exploited to create adversarial outcomes.

The field has seen several key trends emerge: 
\begin{itemize}
    \item increased focus on semantic coherence and naturalness in adversarial examples.
    \item exploitation of powerful generative models to navigate complex latent spaces.
    \item development of geometric understanding of latent vulnerabilities.
    \item enhanced transferability across model architectures.
\end{itemize}
These advances have established LSAs as a distinct and powerful class of adversarial techniques, capable of bypassing many defenses that focus exclusively on input-space perturbations.

Despite progress in defense mechanisms like Latent Adversarial Training~\cite{kumari2019harnessing} and boundary-guided approaches~\cite{zhou2019latent}, latent-space vulnerabilities remain challenging to mitigate comprehensively. The continued evolution of this field suggests that understanding and addressing vulnerabilities in feature representations will remain a critical aspect of developing truly robust computer vision systems in the future.
\section{Emerging Topic of Adversarial Attack in Computer Vision}

\subsection{Explore the Robustness Boundary of Authentication Systems}

Authentication frameworks utilizing biological characteristics have undergone remarkable transformation throughout recent years, shifting from conventional pattern recognition techniques to advanced computational intelligence methodologies.
While this progression has substantially improved identification precision and user experience, it simultaneously generates intricate security vulnerabilities that necessitate a fundamental reconsideration of protection strategies for biological identifiers. Central to these vulnerabilities exists what researchers term the non-replaceable attribute dilemma~\cite{jain2008biometric}. The unchangeable quality of physiological identifiers indicates that upon compromise, individuals cannot simply generate new ones as with alphanumeric credentials. Furthermore, these compromised physiological data points enable creation of effective oppositional instances, establishing significant risks to computational intelligence verification systems. This vulnerability has grown substantially more concerning within the computational learning paradigm.

\subsubsection{The Irrevocability Paradox in Biometric Authentication}

A critical security challenge with biometric authentication systems stems from their unchangeable nature, as documented by~\cite{jain2008biometric}. Security experts identify this as the ``irrevocability paradox'': biometric identifiers including facial features, signature patterns, and fingerprint characteristics cannot be altered or replaced when compromised, unlike conventional authentication methods such as tokens or passwords. The unchanging property that provides convenience in biometric systems simultaneously constitutes their principal security weakness.

The concept of cancelable biometrics emerged as a solution to this vulnerability. This approach, introduced in foundational publications by~\cite{ratha2006cancelable,ratha2007generating}, involves applying systematic and reproducible alterations to biometric information through various mathematical operations. These included functional, polar, and Cartesian transformations applied to fingerprint data points. Instead of storing original biometric information, systems could retain the modified template, allowing for replacement with differently transformed versions if security became compromised. This methodology sought to add renewability while maintaining authentication performance.

The evolution of cancelable biometric techniques continued with the work of~\cite{boult2007revocable}, who created revocable biotokens for fingerprints using binary field encoding strategies that simultaneously improved security aspects and recognition capabilities. Additional advancements came from~\cite{jin2010revocable} through the development of registration-independent fingerprint templates that prioritized template diversity and mathematical non-reversibility. A limitation of these approaches was their concentration on fixed security characteristics rather than resilience against developing attack methodologies.

\subsubsection{Limitations of Static Revocation Mechanisms}

The theoretical foundations of cancelable biometric systems have not translated effectively into practical applications. Research conducted by scholars~\cite{jain2008biometric} reveals substantial operational constraints involving unavoidable compromises among security, discriminability, and computational efficiency. A particularly significant flaw exists in their architectural design, which primarily addresses fixed-state threat models incapable of responding to the evolving nature of contemporary security challenges.

Such architectural limitations became increasingly evident as sophisticated adversarial methodologies developed. Academic literature~\cite{biggio2015adversarial,guo2024white} offers a thorough analysis of biometric authentication frameworks within the context of adversarial machine learning principles. This research identifies critical vulnerabilities including authentication data corruption, synthetic identity presentation, and the substantial challenge of maintaining learning integrity within changeable threat landscapes. Their findings emphasize why protection mechanisms lacking adaptability invariably prove insufficient against strategically evolving security threats.

Further evidence supporting the insufficiency of non-adaptive cancellation protocols emerged through research~\cite{lovisotto2020biometric} that introduced the concept of ``biometric backdoors'' as a specialized data poisoning technique targeting facial verification systems. By imperceptibly influencing the template modification protocols through strategically constructed adversarial inputs, researchers achieved sustained identity falsification capabilities that circumvented established protection measures. This vulnerability demonstrates how systems designed for continuous improvement through adaptation can unintentionally create novel security weaknesses when adversarial considerations are not fully integrated into their design architecture.

\subsubsection{Machine Learning Era: Amplification of Vulnerabilities}

The evolution toward machine learning technologies for biometric authentication, specifically systems utilizing deep learning structures, has considerably magnified challenges related to irrevocability. While providing enhanced recognition capabilities, these technologies have simultaneously generated novel security vulnerabilities which traditional protective frameworks cannot effectively counter.

Research published in~\cite{won2021generative} illustrates how generative adversarial networks successfully fabricate synthetic fingerprint patterns capable of deceiving deep learning recognition mechanisms. Complementary investigations documented in~\cite{marrone2019adversarial} demonstrate that adversarial perturbations, consisting of minimal alterations invisible to human examination, consistently compromise fingerprint authentication protocols. These discoveries underscore the substantial vulnerability of learning algorithms when confronted with sophisticated adversarial strategies.

Within handwritten signature authentication contexts, comprehensive analysis documented in~\cite{hafemann2019characterizing} examined adversarial examples targeting offline signature verification frameworks. This research established that verification systems could be manipulated to incorrectly reject legitimate signatures or accept fabricated ones through deliberately constructed perturbations. Their analysis indicated that lower quality signature specimens demonstrated heightened susceptibility to adversarial manipulation, with error percentages escalating by approximately 49.19\% under specific testing parameters. This weakness presents particular concerns for signature verification technologies since signature data inherently displays significant natural variation and inconsistency.

\nocite{liu2024mmad,zhao2024weak,li2025fedkd,lei2025towards,yam2025my,zhao2025affective}

\subsubsection{Dynamic Adversarial Environments and Adaptive Threats}

Biometric security systems face significant challenges due to the continually transforming landscape of adversarial threats. Research~\cite{poh2012critical} examining adaptive biometric frameworks reveals a crucial conflict between adaptation mechanisms designed to enhance recognition performance and those implemented for security purposes. When biometric systems are engineered to refresh templates for accommodating natural biometric variations, they may unintentionally create vulnerabilities that adaptive adversaries can systematically exploit.

This vulnerability has been substantiated through empirical research. A methodological approach for assessing risks associated with adversarial attacks on biometric technologies~\cite{park2024comprehensive} illustrates how advanced attackers can leverage the relationships between various components within biometric architectures. Such exploitation creates sequential effects that conventional security assessments typically fail to identify. The investigation particularly emphasizes how template adaptation mechanisms remain susceptible to systematic adversarial manipulation across extended periods.

In response to increasingly complex security challenges, the scientific community has begun investigating more flexible defensive strategies. Recent contributions include a contextually aware framework for liveness verification in facial recognition systems~\cite{lavens2023mitigating} that incorporates situational data and surrounding conditions to dynamically adjust security configurations. This research constitutes significant progress toward developing biometric systems capable of adapting to emerging threat patterns, moving beyond traditional static protective measures.

\subsubsection{Challenges of Low-Quality Samples in Signature Verification}

Within the domain of biometric authentication systems, handwritten signatures constitute a particularly problematic category owing to their substantial variance and quality-dependent security implications. Contrary to physiological identifiers that maintain consistency, behavioral authentication factors such as handwritten signatures demonstrate natural fluctuations that simultaneously complicate recognition processes and security protocols.

Research conducted by investigators~\cite{islam2017increasing} illuminated these complexities, revealing that signature inconsistency functions as both advantage and liability: individuals can deliberately alter their signing patterns, providing natural replacement capabilities, while simultaneously creating openings for malicious exploitation of this inherent variability. The investigation highlighted how suboptimal signature acquisition substantially elevates false acceptance probabilities, thereby establishing security vulnerabilities that adversaries can leverage.

Subsequent quantitative analysis~\cite{hafemann2019characterizing} further substantiated these concerns by revealing that antagonistic interventions against signature verification frameworks achieved notably greater success when utilizing or targeting signatures of inferior quality. Experimental evidence demonstrated that verification systems developed using pristine samples remained susceptible when processing degraded inputs, a situation frequently encountered in practical applications. Additionally, conventional protective strategies including adversarial augmentation displayed insufficient efficacy in addressing these vulnerabilities, underscoring requirements for more refined and responsive defensive methodologies.

\subsubsection{Multimodal and Adaptive Defensive Strategies}

The field has witnessed a significant shift toward multimodal biometric frameworks and dynamic protection mechanisms in addressing these security issues. Research contributions~\cite{walia2019adaptive} have introduced an adaptive weighted graph methodology for creating multimodal cancelable biometric templates, illustrating how combining features at fundamental levels can simultaneously strengthen security and maintain template revocability. This methodology intelligently assigns weights to various biometric identifiers according to their quality assessments, thus offering resilience when sample quality fluctuates.

Contemporary studies~\cite{alghamdi2024enhancing} have performed extensive evaluations regarding how various integration techniques influence multimodal biometric system resilience when subjected to adversarial manipulation. Evidence suggests that integration at the input level typically offers greater protection compared to combining information at scoring or decision stages, although optimal integration strategies remain context-dependent. Significantly, this research emphasizes the necessity of incorporating adversarial resilience considerations during the initial system architecture phase rather than implementing protective measures retrospectively.

Concurrently, investigations~\cite{lee2024adversarial} have explored the varying susceptibility levels across different biometric indicators when confronted with adversarial attacks. Research indicates certain identifiers (palmprints specifically) demonstrate heightened vulnerability compared to others (such as iris patterns). These observations suggest potential advantages in strategically selecting and differentially weighting biometric modalities to enhance overall system protection against adversarial threats.

\subsubsection{The Expanding Robustness Boundary and Future Direction}

Contemporary security systems continue to grapple with fundamental challenges despite advancements in countering adversarial threats and addressing irrevocability issues. The research community has yet to fully reconcile the static architecture of current revocation solutions with adversaries' evolutionary capabilities. Studies\cite{park2024comprehensive,lovisotto2020biometric} reveal how attackers continuously refine their methodologies, often circumventing the very protective mechanisms implemented to safeguard systems.

This fundamental incongruity becomes particularly evident when examining machine learning-powered signature authentication frameworks. These systems possess intricate model structures combined with inherent biometric variability, creating extensive vulnerability surfaces that remain insufficiently investigated. Conventional security assessment techniques predominantly examine fixed threat vectors, offering limited protection guarantees within this complex domain, thus necessitating more holistic and adaptable evaluation methodologies.

The scientific community must conduct thorough adversarial boundary testing to strengthen modern biometric frameworks. Through implementation of advanced attack protocols that evaluate systems under realistic and adaptive threat scenarios, researchers can uncover critical weaknesses and subsequently formulate more resilient protection mechanisms. This perspective conceptualizes security as a continuous refinement process rather than a fixed attribute—a consideration especially vital for signature verification technologies where authentication failures carry significant consequences.

As authentication systems incorporating machine learning continue their technological progression, we urgently require innovative frameworks capable of resolving the irrevocability paradox within evolving adversarial landscapes. Such advancement demands not only technical progress in cancelable biometrics and defensive countermeasures but also a paradigm shift in biometric security conceptualization—transitioning from static security assertions toward dynamic assurance frameworks capable of adapting at pace with emerging threats.
Chapter~3 addresses these challenges by introducing our foundational framework for dynamic adversarial evaluation, providing adaptive testing methodologies that evolve alongside threat landscapes and establish the technical foundation for resilient biometric authentication systems.

\subsection{Adversarial Protection against Malicious Generative Vision Models}

The emergence and proliferation of sophisticated visual generative technologies have revolutionized our capacity to synthesize, modify, and create images. These technological innovations, while impressive in their capabilities, concurrently raise substantial concerns regarding security vulnerabilities and potential privacy violations. Such issues encompass unauthorized utilization of individual photographs, violations of copyright protections, and deliberate manipulation of visual media. This segment investigates the chronological progression of protective adversarial strategies designed to counteract unauthorized processing operations conducted by visual generative frameworks. Particular attention is directed toward models based on diffusion principles, given their current preeminence within the landscape of vision-oriented generative architectures.

\subsubsection{Early Defenses Against Diffusion Models}

During 2022 to 2023, with the increasing popularity of diffusion models, academic investigations began to explore adversarial methodologies for protecting visual content from unauthorized utilization. Among the initial contributions in this research area was the work presented in~\cite{advdm}, which introduced AdvDM, a novel adversarial protection framework specifically engineered for diffusion architectures. This approach concentrated on interfering with the reverse diffusion sequence through strategic perturbations of latent representations, consequently preventing unauthorized extraction and replication of artistic elements. The significance of this research lies in its departure from conventional adversarial strategies, as it specifically targeted the sequential denoising operations fundamental to diffusion model functionality.

Subsequent research expanded this conceptual foundation with the introduction of MIST in~\cite{mist}, a sophisticated framework that substantially enhanced cross-model transferability of adversarial perturbations against diffusion systems. The authors of this study refined the adversarial objective function to ensure protective modifications remained robust across varied model structures and implementation parameters. Experimental validation confirmed that systematically constructed adversarial examples could effectively safeguard visual content from unauthorized synthesis operations while preserving perceptual quality.

In parallel developments, research documented in~\cite{zhao2024unlearnable} introduced the Unlearnable Diffusion Perturbation (UDP) methodology, which represented a conceptual shift by focusing on preventing diffusion systems from effectively utilizing protected data during model training. This approach transitioned from inference-phase protection to training-phase safeguards, thereby addressing more comprehensive privacy and intellectual property considerations. The UDP technique implemented imperceptible adversarial modifications to visual content, rendering such materials unsuitable for training generative models such as Stable Diffusion, thus providing protection for creative styles and personal information against unauthorized appropriation.

\subsubsection{Personalization and Identity Protection}

The emergence of techniques for personalized text-to-image generation, exemplified by systems such as DreamBooth~\cite{ruiz2023dreambooth}, has generated significant concerns regarding unauthorized utilization and identity theft. Research presented in~\cite{le2023anti} introduced Anti-DreamBooth, a protective framework that impairs personalized content generation through the application of adversarial perturbations to personal images. This system specifically targets the fundamental learning mechanisms that allow DreamBooth and comparable methods to establish connections between textual descriptions and visual identities, thus inhibiting unauthorized personalized content synthesis.

A different investigation documented in~\cite{liu2024metacloak} established MetaCloak, a framework utilizing meta-learning principles to safeguard against unauthorized subject-based text-to-image generation. This approach generates resilient adversarial disturbances that maintain their effectiveness despite common image transformations including filtering operations and dimensional adjustments. The implementation of meta-learning enables optimization of protective perturbations across various generative models and textual prompt variations, constituting a notable improvement in defense mechanism durability.

The work presented in~\cite{wang2024simac} introduced SimAC, which offers a straightforward approach to preventing personalization, specifically developed for facial privacy protection. This method incorporates time-step-sensitive adversarial noise coupled with feature-based optimization techniques to interrupt identity extraction processes in diffusion models. SimAC achieves an optimal balance between protective efficacy and retention of visual quality by simultaneously targeting frequency domains and encoder layers, rendering it particularly appropriate for protecting portrait imagery.

Addressing the need for instantaneous identity protection, the research in~\cite{guo2024real} established Real-time Identity Defense (RID), a system generating protective perturbations through a single-forward-pass neural network architecture. This efficiency-oriented solution facilitates practical implementation for extensive image protection scenarios, particularly on social media platforms requiring immediate processing capabilities. The RID system demonstrates that effective adversarial protection measures can be implemented without excessive computational requirements, thus addressing a major obstacle to widespread adoption of such protective technologies.

\subsubsection{Robust and Transferable Protection}

Research into enhanced adversarial defense mechanisms emerged following discoveries that initial defenses exhibited insufficient cross-model transferability and susceptibility to preprocessing manipulations. The introduction of Prompt-Agnostic Perturbations (PAP) by researchers~\cite{wan2024prompt} represented a substantial advancement in this domain. This approach utilized Laplace Approximation to conceptualize prompt distributions, generating robust perturbations that maintained efficacy across diverse contexts and prompt configurations. The PAP methodology substantially enhanced the cross-model applicability of adversarial protections, effectively addressing a fundamental weakness inherent in previous techniques.

Protection through Score Distillation Sampling was investigated in scholarly work~\cite{sds}, with particular focus on targeting latent diffusion models' encoder components to establish effective adversarial safeguards. This strategy achieved computational efficiency while sustaining resistance against unauthorized utilization. Concurrent research~\cite{wu2023towards} established an Adversarial Decoupling Augmentation Framework (ADAF) that implemented text-associated augmentations, creating consistent protections against various malicious input prompts, particularly beneficial for facial privacy preservation applications.

Addressing the critical issue of maintaining defensive integrity against preprocessing techniques and adversarial purification methods, investigators~\cite{ozden2024optimization} formulated DiffVax, an optimization-independent framework designed to shield images from unauthorized diffusion-based modifications. This innovation facilitated instantaneous, scalable image protection that retained effectiveness despite preprocessing operations such as compression and dimensional alterations, constituting a noteworthy advancement in practical adversarial defense implementation.

\subsubsection{Hybrid Approaches and Watermarking}

In response to the constraints of purely adversarial techniques, the research community shifted toward integrated solutions that merge adversarial modifications with additional safeguarding mechanisms. Research presented in~\cite{zhu2024watermark} established a framework for Watermark-embedded Adversarial Examples, utilizing conditional generative adversarial networks to create adversarial samples that compel diffusion models to generate outputs containing visible watermarks. This integrated protection framework enables simultaneous copyright verification and quality reduction, tackling multiple dimensions of unauthorized utilization concurrently.

In a parallel research direction, scholars in~\cite{tan2023somewhat} formulated Robust Invisible Watermarking (RIW), a method leveraging adversarial principles to ensure watermark persistence throughout transformations based on diffusion processes. Their methodology maintained extraction accuracy rates of 96\% following content modification, substantiating the effectiveness of permanent identity integration as a supplementary protective strategy.

\subsubsection{Evaluation and Limitations}

The assessment of protective measures' efficacy and constraints has evolved into a systematic research focus as this domain advanced. A framework known as IMPRESS was established by researchers~\cite{cao2023impress} to assess how resilient imperceptible perturbations are against unauthorized utilization of data in generative AI based on diffusion models. Their investigation uncovered that current methodologies remain susceptible to purification approaches, which accentuates the persistent difficulty in developing genuinely robust protective systems.

Further investigations have raised significant concerns regarding the capability of protective perturbations to shield personal information from exploitation. Research conducted on Stable Diffusion~\cite{zhao2024can} introduced the purification methodology GrIDPure, which demonstrates the capacity to circumvent adversarial protections while maintaining image utility, thus exposing fundamental weaknesses in contemporary defensive techniques. Comparable findings were reported in additional studies~\cite{an2024rethinking}, which illustrated how techniques focusing on feature alignment, such as INSIGHT, could effectively neutralize protective perturbations that are imperceptible, thereby enabling diffusion models to regenerate features from images that were supposedly protected. These research outcomes emphasize the continuous evolution within the defense-attack ecosystem and highlight the requirement for protection mechanisms that adapt progressively.

\subsubsection{Recent Advancements and Specialized Defenses}

The literature reveals an evolution toward context-specific protective mechanisms tailored to unique security threats and application environments. Research presented in~\cite{choi2025diffusionguard} established DiffusionGuard as a defensive framework that counters unauthorized diffusion-based image manipulation by focusing on initial diffusion phases while implementing mask-augmentation techniques. This innovation specifically tackles the prevention of illicit partial content alterations, addressing a significant security vulnerability in real-world implementations.

A methodology termed Dual-Domain Anti-Personalization (DDAP) was formulated in~\cite{yang2024ddap}, introducing concurrent disturbances across both spatial and frequency realms to impede texture and detail acquisition in customized text-to-image generation systems. The simultaneous application across multiple domains enables DDAP to attain enhanced resilience in comparison with conventional single-domain techniques, thus validating the efficacy of comprehensive protection frameworks.

Concerning the targeted safeguarding of visual elements, the VCPro framework outlined in~\cite{mi2024visual} implements strategic adversarial modifications to combat counterfeit image creation and stylistic reproduction while maintaining fundamental perceptual integrity. This nuanced protection strategy facilitates adaptable implementation contexts wherein security requirements focus exclusively on particular components within visual content.

\subsubsection{Critical Gaps and Future Directions}

Although considerable advancements have occurred in developing protective measures against adversarial attacks on diffusion models, numerous important research challenges remain unexplored. A significant oversight pertains to neural style transfer applications, which pose unique threats to artists' intellectual property rights. While substantial research addresses generalized image manipulation and personalized content generation, the scholarly community has insufficiently investigated protective techniques specifically tailored for style transfer scenarios. This gap is directly addressed in Chapter~4, where we develop targeted protective techniques specifically for neural style transfer scenarios, providing a comprehensive protection framework for artistic intellectual property.

An additional limitation involves computational efficiency and knowledge requirements. Contemporary adversarial techniques targeting diffusion frameworks typically necessitate comprehensive understanding of model internals, particularly regarding UNet architectural specifications. This requirement substantially restricts cross-model applicability and imposes excessive computational burdens, rendering practical implementation challenging. The field would benefit significantly from methodologies maintaining effectiveness while requiring minimal model information. These limitations are tackled through our grey-box framework in Chapter~5, which requires minimal model knowledge while maintaining effectiveness.

These unaddressed challenges represent fertile ground for subsequent scholarly investigation, potentially yielding more comprehensive and implementable safeguards against unauthorized utilization of visual assets by generative AI systems.
\section{Conclusion}
\subsection{Summary of Key Findings}
This comprehensive literature review has examined the evolving landscape of adversarial attacks in computer vision systems, revealing significant developments across multiple attack paradigms and application domains. Our analysis demonstrates that adversarial vulnerabilities represent both fundamental challenges to the reliability of deep learning systems and valuable tools for security assessment and protection.

\paragraph{Methodological Evolution:} The field has progressed from simple gradient-based perturbations (FGSM, PGD) to sophisticated optimization techniques incorporating momentum, adaptive step sizes, and advanced transferability mechanisms. Pixel-space attacks have evolved beyond basic sign-function approaches to embrace raw gradient methods and specialized optimization frameworks, significantly improving both effectiveness and cross-model transferability.

\paragraph{Physical World Implications:} Physically realizable attacks have successfully bridged the gap between digital vulnerabilities and real-world threats. The development from basic adversarial patches to sophisticated 3D textures, wearable attacks, and dynamic optical perturbations demonstrates the practical severity of adversarial vulnerabilities in deployed systems.

\paragraph{Latent-Space Sophistication:} Latent-space attacks have emerged as a particularly powerful paradigm, leveraging the semantic structure of internal representations to create more transferable and semantically meaningful adversarial examples. The progression from simple VAE manipulations to sophisticated diffusion-model-based attacks reflects the field's adaptation to advancing generative technologies.

\paragraph{Dual Nature of Applications:} Perhaps most significantly, our review reveals the dual nature of adversarial techniques. While these methods expose critical vulnerabilities in computer vision systems, they also serve as valuable tools for robustness assessment and protective applications, including biometric security evaluation and prevention of unauthorized content generation.

\subsection{Challenges and Future Directions}

Building upon the significant contributions outlined in the previous sections, several critical challenges and potential directions for future research in adversarial machine learning for computer vision remain.

\subsubsection{Enhancing Transferability in Adversarial Attacks}

One of the persistent challenges in adversarial machine learning is the transferability of attacks across different model architectures and domains. While recent work has demonstrated promising results in specific contexts, such as the high transferability of attacks across various Latent Diffusion Models, several issues remain unresolved:

\begin{itemize}
\item \textbf{Cross-architecture transferability}:
The effectiveness of adversarial attacks often diminishes when applied to models with fundamentally different architectures~\cite{mopuri2017fast}. Future research should focus on identifying and exploiting common vulnerabilities across diverse neural network designs, potentially through more abstract representations of model behavior rather than architecture-specific features.

\item \textbf{Domain adaptation for adversarial attacks}:
Developing techniques that can adapt adversarial perturbations to new domains without requiring extensive modification would significantly enhance the practical utility of these methods. This could involve learning domain-invariant features~\cite{lu2022domaininvariant} that are consistently susceptible to perturbation across different contexts.
\item \textbf{Model-agnostic protection mechanisms}:
As demonstrated by the Posterior Collapse Attack, targeting fundamental components common across multiple model architectures (such as Variational Autoencoders (VAE) in LDMs) offers a promising direction for creating more universal protection mechanisms. Extending this approach to other shared components or principles in machine learning models represents an important avenue for future work.
\end{itemize}

\subsubsection{Balancing Perturbation Imperceptibility and Attack Effectiveness}

Throughout our research, a consistent challenge has been striking the optimal balance between the imperceptibility of adversarial perturbations and their effectiveness in achieving the desired outcome. This trade-off varies across different applications:
\begin{itemize}
\item \textbf{Domain-adaptive perturbation constraints}: Future research should explore methods for dynamically adjusting perturbation constraints based on image content, viewing conditions, and application context. For instance, images with high-frequency textures may tolerate larger perturbations without perceptible changes compared to smoother images.
\item \textbf{Perceptually guided adversarial optimization}: Incorporating more sophisticated models of human visual perception into adversarial optimization processes could lead to perturbations that more effectively exploit the gaps between machine and human vision. This may include leveraging insights from psychophysics to identify perturbations that are imperceptible to humans while maximally disrupting machine learning models.
\end{itemize}

\subsubsection{Extending to Emerging AI Paradigms}

As AI continues to evolve, new paradigms and architectures emerge, presenting both challenges and opportunities for adversarial machine learning:
\begin{itemize}
\item \textbf{Foundation models and large vision models}: The rise of foundation models~\cite{radford2021learning,oquabdinov2} and large vision models~\cite{bai2024sequential} introduces new vulnerabilities and protection requirements. Exploring adversarial attacks and defenses in these contexts, particularly in multi-modal systems that combine vision and language, represents an important direction for future research.
\item \textbf{Adversarial robustness in self-supervised learning}: As self-supervised learning becomes more prevalent in computer vision, understanding the unique vulnerabilities and robustness properties of these approaches compared to supervised learning is crucial for developing effective protection mechanisms.
\item \textbf{Quantum machine learning}: Looking further ahead, the emergence of quantum machine learning algorithms~\cite{biamonte2017quantum} may fundamentally change the landscape of adversarial attacks and defenses. Investigating the implications of quantum computing for adversarial robustness represents a long-term research direction.
\end{itemize}

\subsection{Implications and Future Outlook}

The findings of this review have several important implications for the computer vision community. First, the rapid evolution of attack methodologies demonstrates that adversarial robustness cannot be treated as a static property but requires continuous adaptation to emerging threats. Second, the success of physically realizable attacks underscores the critical need for robustness evaluation under real-world conditions rather than purely digital settings.

The emerging applications of adversarial techniques for protective purposes represent a paradigm shift in how we conceptualize these methods. Rather than viewing adversarial attacks solely as security threats, the community is increasingly recognizing their potential as tools for privacy protection and intellectual property safeguarding.

Future research should prioritize the development of more generalized attack and defense frameworks that can adapt to rapidly evolving AI technologies. This includes investigating the fundamental principles underlying adversarial vulnerabilities and developing theoretical frameworks that can guide both attack development and defense strategies.

The field would benefit from increased collaboration between adversarial ML researchers and domain experts in critical applications such as healthcare, autonomous systems, and security. Such interdisciplinary approaches could lead to more practical and effective solutions that address real-world deployment challenges.

As computer vision systems become increasingly integrated into society's critical infrastructure, ensuring their adversarial robustness is not merely an academic pursuit but a societal imperative. The continued advancement of both attack and defense methodologies will be essential for building trustworthy AI systems that can operate reliably in adversarial environments.

\bibliographystyle{unsrt}
\bibliography{references}

\begin{thebibliography}{100}

\bibitem{jia2025uni}
Yanhao Jia, Xinyi Wu, Hao Li, Qinglin Zhang, Yuxiao Hu, Shuai Zhao, and Wenqi Fan.
\newblock Uni-retrieval: A multi-style retrieval framework for stem's education.
\newblock {\em arXiv preprint arXiv:2502.05863}, 2025.

\bibitem{jia2025towards}
Yanhao Jia, Xinyi Wu, Qinglin Zhang, Yiran Qin, Luwei Xiao, and Shuai Zhao.
\newblock Towards robust evaluation of stem education: Leveraging mllms in project-based learning.
\newblock {\em arXiv preprint arXiv:2505.17050}, 2025.

\bibitem{guo2023siamese}
Zhongliang Guo, Ognjen Arandjelovi{\'c}, David Reid, Yaxiong Lei, and Jochen B{\"u}ttner.
\newblock A siamese transformer network for zero-shot ancient coin classification.
\newblock {\em Journal of Imaging}, 9(6):107, 2023.

\bibitem{szegedy2013intriguing}
Christian Szegedy, Wojciech Zaremba, Ilya Sutskever, Joan Bruna, Dumitru Erhan, Ian Goodfellow, and Rob Fergus.
\newblock Intriguing properties of neural networks.
\newblock In {\em International Conference on Learning Representations}, 2014.

\bibitem{goodfellow2014explaining}
Ian~J Goodfellow, Jonathon Shlens, and Christian Szegedy.
\newblock Explaining and harnessing adversarial examples.
\newblock In {\em International Conference on Learning Representations}, 2015.

\bibitem{simonyan2014very}
K~Simonyan and A~Zisserman.
\newblock Very deep convolutional networks for large-scale image recognition.
\newblock In {\em International Conference on Learning Representations}, 2015.

\bibitem{verdoliva2020media}
Luisa Verdoliva.
\newblock Media forensics and deepfakes: an overview.
\newblock {\em IEEE Journal of Selected Topics in Signal Processing}, 14(5):910--932, 2020.

\bibitem{guo2024grey}
Zhongliang Guo, Chun~Tong Lei, Lei Fang, Shuai Zhao, Yifei Qian, Jingyu Lin, Zeyu Wang, Cunjian Chen, Ognjen Arandjelovi{\'c}, and Chun~Pong Lau.
\newblock A grey-box attack against latent diffusion model-based image editing by posterior collapse.
\newblock {\em arXiv preprint arXiv:2408.10901}, 2024.

\bibitem{guo2025building}
Zhongliang Guo.
\newblock {\em Building trustworthy computer vision: adversarial techniques for robustness assessment and misuse prevention}.
\newblock PhD thesis, The University of St Andrews, 2025.

\bibitem{guo2024artwork}
Zhongliang Guo, Junhao Dong, Yifei Qian, Kaixuan Wang, Weiye Li, Ziheng Guo, Yuheng Wang, Yanli Li, Ognjen Arandjelovi{\'c}, and Lei Fang.
\newblock Artwork protection against neural style transfer using locally adaptive adversarial color attack.
\newblock In {\em ECAI 2024}, pages 1414--1421. IOS Press, 2024.

\bibitem{guo2025artwork}
Zhongliang Guo, Yifei Qian, Shuai Zhao, Junhao Dong, Yanli Li, Ognjen Arandjelovi{\'c}, Lei Fang, and Chun~Pong Lau.
\newblock Artwork protection against unauthorized neural style transfer and aesthetic color distance metric.
\newblock {\em Pattern Recognition}, page 112105, 2025.

\bibitem{eykholt2018robust}
Kevin Eykholt, Ivan Evtimov, Earlence Fernandes, Bo~Li, Amir Rahmati, Chaowei Xiao, Atul Prakash, Tadayoshi Kohno, and Dawn Song.
\newblock Robust physical-world attacks on deep learning visual classification.
\newblock In {\em Proceedings of the IEEE Conference on Computer Vision and Pattern Recognition (CVPR)}, 2018.

\bibitem{finlayson2018adversarial}
Samuel~G Finlayson, Hyung~Won Chung, Isaac~S Kohane, and Andrew~L Beam.
\newblock Adversarial attacks against medical deep learning systems.
\newblock {\em arXiv preprint arXiv:1804.05296}, 2018.

\bibitem{cartella2021adversarial}
Francesco Cartella, Orlando Anunciacao, Yuki Funabiki, Daisuke Yamaguchi, Toru Akishita, and Olivier Elshocht.
\newblock Adversarial attacks for tabular data: Application to fraud detection and imbalanced data.
\newblock {\em arXiv preprint arXiv:2101.08030}, 2021.

\bibitem{zhang2019adversarial}
Jiliang Zhang and Chen Li.
\newblock Adversarial examples: Opportunities and challenges.
\newblock {\em IEEE transactions on neural networks and learning systems}, 31(7):2578--2593, 2019.

\bibitem{yuan2019adversarial}
Xiaoyong Yuan, Pan He, Qile Zhu, and Xiaolin Li.
\newblock Adversarial examples: Attacks and defenses for deep learning.
\newblock {\em IEEE transactions on neural networks and learning systems}, 30(9):2805--2824, 2019.

\bibitem{shan2020fawkes}
Shawn Shan, Emily Wenger, Jiayun Zhang, Huiying Li, Haitao Zheng, and Ben~Y Zhao.
\newblock Fawkes: Protecting privacy against unauthorized deep learning models.
\newblock In {\em 29th USENIX security symposium (USENIX Security 20)}, pages 1589--1604, 2020.

\bibitem{adi2018turning}
Yossi Adi, Carsten Baum, Moustapha Cisse, Benny Pinkas, and Joseph Keshet.
\newblock Turning your weakness into a strength: Watermarking deep neural networks by backdooring.
\newblock In {\em 27th USENIX security symposium (USENIX Security 18)}, pages 1615--1631, 2018.

\bibitem{madry2018towards}
Aleksander Madry, Aleksandar Makelov, Ludwig Schmidt, Dimitris Tsipras, and Adrian Vladu.
\newblock Towards deep learning models resistant to adversarial attacks.
\newblock In {\em International Conference on Learning Representations}, 2018.

\bibitem{yuan2022natural}
Shengming Yuan, Qilong Zhang, Lianli Gao, Yaya Cheng, and Jingkuan Song.
\newblock Natural color fool: Towards boosting black-box unrestricted attacks.
\newblock In {\em Advances in Neural Information Processing Systems}, volume~35, pages 7546--7560, 2022.

\bibitem{liu2023instruct2attack}
Jiang Liu, Chen Wei, Yuxiang Guo, Heng Yu, Alan Yuille, Soheil Feizi, Chun~Pong Lau, and Rama Chellappa.
\newblock {Instruct2Attack}: Language-guided semantic adversarial attacks.
\newblock {\em arXiv preprint arXiv:2311.15551}, 2023.

\bibitem{xu2020adversarial}
Han Xu, Yao Ma, Hao-Chen Liu, Debayan Deb, Hui Liu, Ji-Liang Tang, and Anil~K Jain.
\newblock Adversarial attacks and defenses in images, graphs and text: A review.
\newblock {\em International Journal of Automation and Computing}, 17:151--178, 2020.

\bibitem{qiu2019review}
Shilin Qiu, Qihe Liu, Shijie Zhou, and Chunjiang Wu.
\newblock Review of artificial intelligence adversarial attack and defense technologies.
\newblock {\em Applied Sciences}, 9(5):909, 2019.

\bibitem{dong2018boosting}
Yinpeng Dong, Fangzhou Liao, Tianyu Pang, Hang Su, Jun Zhu, Xiaolin Hu, and Jianguo Li.
\newblock Boosting adversarial attacks with momentum.
\newblock In {\em Proceedings of the IEEE Conference on Computer Vision and Pattern Recognition (CVPR)}, 2018.

\bibitem{zhang2023boosting}
Qikun Zhang, Yuzhi Zhang, Yanling Shao, Mengqi Liu, Jianyong Li, Junling Yuan, and Ruifang Wang.
\newblock Boosting adversarial attacks with nadam optimizer.
\newblock {\em Electronics}, 12(6):1464, 2023.

\bibitem{papernot2017practical}
Nicolas Papernot, Patrick McDaniel, Ian Goodfellow, Somesh Jha, Z~Berkay Celik, and Ananthram Swami.
\newblock Practical black-box attacks against machine learning.
\newblock In {\em Proceedings of the 2017 ACM on Asia conference on computer and communications security}, pages 506--519, 2017.

\bibitem{tramer2018ensemble}
Florian Tram{\`e}r, Alexey Kurakin, Nicolas Papernot, Ian Goodfellow, Dan Boneh, and Patrick McDaniel.
\newblock Ensemble adversarial training: Attacks and defenses.
\newblock In {\em International Conference on Learning Representations}, 2018.

\bibitem{dong2023restricted}
Junhao Dong, Yuan Wang, Jianhuang Lai, and Xiaohua Xie.
\newblock Restricted black-box adversarial attack against deepfake face swapping.
\newblock {\em IEEE Transactions on Information Forensics and Security}, 2023.

\bibitem{andriushchenko2020square}
Maksym Andriushchenko, Francesco Croce, Nicolas Flammarion, and Matthias Hein.
\newblock Square attack: a query-efficient black-box adversarial attack via random search.
\newblock In {\em European Conference on Computer Vision}, pages 484--501. Springer, 2020.

\bibitem{zhang2021progressive}
Jiawei Zhang, Linyi Li, Huichen Li, Xiaolu Zhang, Shuang Yang, and Bo~Li.
\newblock Progressive-scale boundary blackbox attack via projective gradient estimation.
\newblock In {\em International Conference on Machine Learning}, pages 12479--12490. PMLR, 2021.

\bibitem{wang2023lfaa}
Kunyu Wang, Juluan Shi, and Wenxuan Wang.
\newblock {LFAA}: Crafting transferable targeted adversarial examples with low-frequency perturbations.
\newblock In {\em ECAI 2023}, pages 2483--2490. IOS Press, 2023.

\bibitem{liang2022lp}
Qi~Liang, Qiang Li, and Song Yang.
\newblock {LP-GAN}: Learning perturbations based on generative adversarial networks for point cloud adversarial attacks.
\newblock {\em Image and Vision Computing}, 120:104370, 2022.

\bibitem{he2016deep}
Kaiming He, Xiangyu Zhang, Shaoqing Ren, and Jian Sun.
\newblock Deep residual learning for image recognition.
\newblock In {\em Proceedings of the IEEE Conference on Computer Vision and Pattern Recognition (CVPR)}, 2016.

\bibitem{wang2004image}
Zhou Wang, Alan~C Bovik, Hamid~R Sheikh, and Eero~P Simoncelli.
\newblock Image quality assessment: from error visibility to structural similarity.
\newblock {\em IEEE Transactions on Image Processing}, 13(4):600--612, 2004.

\bibitem{heusel2017gans}
Martin Heusel, Hubert Ramsauer, Thomas Unterthiner, Bernhard Nessler, and Sepp Hochreiter.
\newblock {GANs} trained by a two time-scale update rule converge to a local nash equilibrium.
\newblock In {\em Advances in Neural Information Processing Systems}, volume~30, 2017.

\bibitem{zhang2018unreasonable}
Richard Zhang, Phillip Isola, Alexei~A. Efros, Eli Shechtman, and Oliver Wang.
\newblock The unreasonable effectiveness of deep features as a perceptual metric.
\newblock In {\em Proceedings of the IEEE Conference on Computer Vision and Pattern Recognition (CVPR)}, 2018.

\bibitem{kurakin2018adversarial}
Alexey Kurakin, Ian~J Goodfellow, and Samy Bengio.
\newblock Adversarial examples in the physical world.
\newblock In {\em Artificial Intelligence Safety and Security}, pages 99--112. Chapman and Hall/CRC, 2018.

\bibitem{lin2020nesterov}
Jiadong Lin, Chuanbiao Song, Kun He, Liwei Wang, and John~E. Hopcroft.
\newblock Nesterov accelerated gradient and scale invariance for adversarial attacks.
\newblock In {\em International Conference on Learning Representations}, 2020.

\bibitem{cheng2021fast}
Yaya Cheng, Jingkuan Song, Xiaosu Zhu, Qilong Zhang, Lianli Gao, and Heng~Tao Shen.
\newblock Fast gradient non-sign methods.
\newblock {\em arXiv preprint arXiv:2110.12734}, 2021.

\bibitem{yuan2024adaptive}
Zheng Yuan, Jie Zhang, Zhaoyan Jiang, Liangliang Li, and Shiguang Shan.
\newblock Adaptive perturbation for adversarial attack.
\newblock {\em IEEE Transactions on Pattern Analysis and Machine Intelligence}, 2024.

\bibitem{yang2023rethinking}
Junjie Yang, Tianlong Chen, Xuxi Chen, Zhangyang Wang, and Yingbin Liang.
\newblock Rethinking {PGD} attack: Is sign function necessary?
\newblock {\em CoRR}, 2023.

\bibitem{wang2021generalizing}
Yixiang Wang, Jiqiang Liu, and Xiaolin Chang.
\newblock Generalizing adversarial examples by adabelief optimizer.
\newblock {\em arXiv preprint arXiv:2101.09930}, 2021.

\bibitem{zhou2024study}
Ronghui Zhou.
\newblock Study of optimiser-based enhancement of adversarial attacks on neural networks.
\newblock In {\em 2024 International Conference on Interactive Intelligent Systems and Techniques (IIST)}, pages 740--747, 2024.

\bibitem{tao2023adapting}
Wei Tao, Lei Bao, Sheng Long, Gaowei Wu, and Qing Tao.
\newblock Adapting step-size: A unified perspective to analyze and improve gradient-based methods for adversarial attacks.
\newblock {\em arXiv preprint arXiv:2301.11546}, 2023.

\bibitem{long2024convergence}
Sheng Long, Wei Tao, LI~Shuohao, Jun Lei, and Jun Zhang.
\newblock On the convergence of an adaptive momentum method for adversarial attacks.
\newblock {\em Proceedings of the AAAI Conference on Artificial Intelligence}, 38(13):14132--14140, 2024.

\bibitem{huang2024using}
Qidong Huang, Leiji Lu, Jun Chen, and Lei Bao.
\newblock Using dynamically changing step sizes to increase the success rate of adversarial attacks.
\newblock In {\em 2024 5th International Conference on Computer Vision, Image and Deep Learning (CVIDL)}, pages 1071--1076. IEEE, 2024.

\bibitem{chiang2020witchcraft}
Ping-Yeh Chiang, Jonas Geiping, Micah Goldblum, Tom Goldstein, Renkun Ni, Steven Reich, and Ali Shafahi.
\newblock {WITCHcraft}: Efficient pgd attacks with random step size.
\newblock In {\em ICASSP 2020-2020 IEEE International Conference on Acoustics, Speech and Signal Processing (ICASSP)}, pages 3747--3751. IEEE, 2020.

\bibitem{wang2022enhancing}
Guoqiu Wang, Huanqian Yan, and Xingxing Wei.
\newblock Enhancing transferability of adversarial examples with spatial momentum.
\newblock In {\em Chinese Conference on Pattern Recognition and Computer Vision (PRCV)}, pages 593--604. Springer, 2022.

\bibitem{zhang2023improving}
Zeliang Zhang, Peihan Liu, Xiaosen Wang, and Chenliang Xu.
\newblock Improving adversarial transferability with scheduled step size and dual example.
\newblock {\em arXiv preprint arXiv:2301.12968}, 2023.

\bibitem{fang2023adversarial}
Youqing Fang, Jingwen Jia, Yuhai Yang, and Wan-Li Lyu.
\newblock Adversarial example generation method based on probability histogram equalization.
\newblock In {\em 2023 42nd Chinese Control Conference (CCC)}, pages 7951--7958, 2023.

\bibitem{jang2022strengthening}
Donggon Jang, Sanghyeok Son, and Dae-Shik Kim.
\newblock Strengthening the transferability of adversarial examples using advanced looking ahead and {Self-CutMix}.
\newblock In {\em Proceedings of the IEEE/CVF Conference on Computer Vision and Pattern Recognition (CVPR) Workshops}, pages 147--154, 2022.

\bibitem{zhu2023ICCV}
Hegui Zhu, Yuchen Ren, Xiaoyan Sui, Lianping Yang, and Wuming Jiang.
\newblock Boosting adversarial transferability via gradient relevance attack.
\newblock In {\em Proceedings of the IEEE/CVF International Conference on Computer Vision (ICCV)}, pages 4741--4750, 2023.

\bibitem{wan2023average}
Chen Wan, Fangjun Huang, and Xianfeng Zhao.
\newblock Average gradient-based adversarial attack.
\newblock {\em IEEE Transactions on Multimedia}, 25:9572--9585, 2023.

\bibitem{liu2020f}
Sijie Liu, Zhixiang Zhang, Xian Zhang, and Haojun Feng.
\newblock {F-MIFGSM}: adversarial attack algorithm for the feature region.
\newblock In {\em 2020 IEEE 9th joint international information technology and artificial intelligence conference (ITAIC)}, volume~9, pages 2164--2170. IEEE, 2020.

\bibitem{xu2022fast}
Zhefeng Xu, Zhijian Luo, and Jinlong Mu.
\newblock Fast gradient scaled method for generating adversarial examples.
\newblock In {\em Proceedings of the 2022 6th International Conference on Innovation in Artificial Intelligence}, pages 189--193, 2022.

\bibitem{xu2022scale}
Mengting Xu, Tao Zhang, Zhongnian Li, and Daoqiang Zhang.
\newblock Scale-invariant adversarial attack for evaluating and enhancing adversarial defenses.
\newblock {\em arXiv preprint arXiv:2201.12527}, 2022.

\bibitem{lad2024fast}
Aarti Lad, Ruchi Bhale, and Shlok Belgamwar.
\newblock Fast gradient sign method ({FGSM}) variants in white box settings: A comparative study.
\newblock In {\em 2024 International Conference on Inventive Computation Technologies (ICICT)}, pages 382--386. IEEE, 2024.

\bibitem{athalye2018synthesizing}
Anish Athalye, Logan Engstrom, Andrew Ilyas, and Kevin Kwok.
\newblock Synthesizing robust adversarial examples.
\newblock In {\em International Conference on Machine Learning}, pages 284--293. PMLR, 2018.

\bibitem{wei2022adversarial}
Xingxing Wei, Ying Guo, and Jie Yu.
\newblock Adversarial sticker: A stealthy attack method in the physical world.
\newblock {\em IEEE Transactions on Pattern Analysis and Machine Intelligence}, 45(3):2711--2725, 2022.

\bibitem{huang2020universal}
Lifeng Huang, Chengying Gao, Yuyin Zhou, Cihang Xie, Alan~L. Yuille, Changqing Zou, and Ning Liu.
\newblock Universal physical camouflage attacks on object detectors.
\newblock In {\em IEEE/CVF Conference on Computer Vision and Pattern Recognition (CVPR)}, 2020.

\bibitem{HU2021ICCV}
Yu-Chih-Tuan Hu, Bo-Han Kung, Daniel~Stanley Tan, Jun-Cheng Chen, Kai-Lung Hua, and Wen-Huang Cheng.
\newblock Naturalistic physical adversarial patch for object detectors.
\newblock In {\em Proceedings of the IEEE/CVF International Conference on Computer Vision (ICCV)}, pages 7848--7857, 2021.

\bibitem{cheng2022physical}
Zhiyuan Cheng, James Liang, Hongjun Choi, Guanhong Tao, Zhiwen Cao, Dongfang Liu, and Xiangyu Zhang.
\newblock Physical attack on monocular depth estimation with optimal adversarial patches.
\newblock In {\em European Conference on Computer Vision}, pages 514--532. Springer, 2022.

\bibitem{yang2023towards}
Xiao Yang, Chang Liu, Longlong Xu, Yikai Wang, Yinpeng Dong, Ning Chen, Hang Su, and Jun Zhu.
\newblock Towards effective adversarial textured {3D} meshes on physical face recognition.
\newblock In {\em Proceedings of the IEEE/CVF Conference on Computer Vision and Pattern Recognition (CVPR)}, pages 4119--4128, 2023.

\bibitem{hu2023CVPR}
Zhanhao Hu, Wenda Chu, Xiaopei Zhu, Hui Zhang, Bo~Zhang, and Xiaolin Hu.
\newblock Physically realizable natural-looking clothing textures evade person detectors via 3d modeling.
\newblock In {\em Proceedings of the IEEE/CVF Conference on Computer Vision and Pattern Recognition (CVPR)}, pages 16975--16984, 2023.

\bibitem{tan2021legitimate}
Jia Tan, Nan Ji, Haidong Xie, and Xueshuang Xiang.
\newblock Legitimate adversarial patches: Evading human eyes and detection models in the physical world.
\newblock In {\em Proceedings of the 29th ACM International Conference on Multimedia}, pages 5307--5315, 2021.

\bibitem{zarei2022adversarial}
Mohammad Zarei, Chris Ward, Josh Harguess, and Marshal Aiken.
\newblock Adversarial barrel! an evaluation of {3D} physical adversarial attacks.
\newblock In {\em 2022 IEEE Applied Imagery Pattern Recognition Workshop (AIPR)}, pages 1--6, 2022.

\bibitem{lovisotto2021slap}
Giulio Lovisotto, Henry Turner, Ivo Sluganovic, Martin Strohmeier, and Ivan Martinovic.
\newblock $\{$SLAP$\}$: Improving physical adversarial examples with $\{$Short-Lived$\}$ adversarial perturbations.
\newblock In {\em 30th USENIX Security Symposium (USENIX Security 21)}, pages 1865--1882, 2021.

\bibitem{han2023don}
Yi~Han, Matthew Chan, Eric Wengrowski, Zhuohuan Li, Nils~Ole Tippenhauer, Mani Srivastava, Saman Zonouz, and Luis Garcia.
\newblock Why don't you clean your glasses? perception attacks with dynamic optical perturbations.
\newblock {\em arXiv preprint arXiv:2307.13131}, 2023.

\bibitem{wang2023ICCV}
Donghua Wang, Wen Yao, Tingsong Jiang, Chao Li, and Xiaoqian Chen.
\newblock {RFLA}: A stealthy reflected light adversarial attack in the physical world.
\newblock In {\em Proceedings of the IEEE/CVF International Conference on Computer Vision (ICCV)}, pages 4455--4465, 2023.

\bibitem{wei2023CVPR}
Xingxing Wei, Jie Yu, and Yao Huang.
\newblock Physically adversarial infrared patches with learnable shapes and locations.
\newblock In {\em Proceedings of the IEEE/CVF Conference on Computer Vision and Pattern Recognition (CVPR)}, pages 12334--12342, 2023.

\bibitem{wang2021dual}
Jiakai Wang, Aishan Liu, Zixin Yin, Shunchang Liu, Shiyu Tang, and Xianglong Liu.
\newblock Dual attention suppression attack: Generate adversarial camouflage in physical world.
\newblock In {\em Proceedings of the IEEE/CVF Conference on Computer Vision and Pattern Recognition (CVPR)}, pages 8565--8574, 2021.

\bibitem{wang2024stealthy}
Ce~Wang and Qianmu Li.
\newblock Stealthy adversarial patch for evading object detectors based on sensitivity maps.
\newblock In {\em 2024 IEEE Cyber Science and Technology Congress (CyberSciTech)}, pages 322--328. IEEE, 2024.

\bibitem{zhao2025local}
Ling Zhao, Xun Lv, Lili Zhu, Binyan Luo, Hang Cao, Jiahao Cui, Haifeng Li, and Jian Peng.
\newblock A local adversarial attack with a maximum aggregated region sparseness strategy for 3d objects.
\newblock {\em Journal of Imaging}, 11(1):25, 2025.

\bibitem{qian2024semi}
Yifei Qian, Xiaopeng Hong, Zhongliang Guo, Ognjen Arandjelovi{\'c}, and Carl~R Donovan.
\newblock Semi-supervised crowd counting with contextual modeling: Facilitating holistic understanding of crowd scenes.
\newblock {\em IEEE Transactions on Circuits and Systems for Video Technology}, 2024.

\bibitem{li2024threats}
Yanli Li, Jifei Hu, Zhongliang Guo, Nan Yang, Huaming Chen, Dong Yuan, and Weiping Ding.
\newblock Threats and defenses in federated learning life cycle: A comprehensive survey and challenges.
\newblock {\em IEEE Transactions on Neural Networks and Learning Systems}, 2025.

\bibitem{qian2025perspective}
Yifei Qian, Liangfei Zhang, Zhongliang Guo, Xiaopeng Hong, Ognjen Arandjelovi{\'c}, and Carl~R Donovan.
\newblock Perspective-assisted prototype-based learning for semi-supervised crowd counting.
\newblock {\em Pattern Recognition}, 158:111073, 2025.

\bibitem{zhao2024survey}
Shuai Zhao, Meihuizi Jia, Zhongliang Guo, Leilei Gan, Xiaoyu Xu, Xiaobao Wu, Jie Fu, Feng Yichao, Fengjun Pan, and Anh~Tuan Luu.
\newblock A survey of recent backdoor attacks and defenses in large language models.
\newblock {\em Transactions on Machine Learning Research}, 2025.
\newblock Survey Certification.

\bibitem{hu2025syntactic}
Man Hu, Yatao Yang, Deng Pan, Zhongliang Guo, Luwei Xiao, Deyu Lin, and Shuai Zhao.
\newblock Syntactic paraphrase-based synthetic data generation for backdoor attacks against chinese language models.
\newblock {\em Information Fusion}, page 103376, 2025.

\bibitem{li2025achieving}
Yuqi Li, Yanli Li, Kai Zhang, Fuyuan Zhang, Chuanguang Yang, Zhongliang Guo, Weiping Ding, and Tingwen Huang.
\newblock Achieving fair medical image segmentation in foundation models with adversarial visual prompt tuning.
\newblock {\em Information Sciences}, page 122501, 2025.

\bibitem{liu2025distillation}
Wei Liu, Yonglin Wu, Chaoqun Li, Zhuodong Liu, and Huanqian Yan.
\newblock Distillation-enhanced physical adversarial attacks.
\newblock {\em arXiv preprint arXiv:2501.02232}, 2025.

\bibitem{lian2022benchmarking}
Jiawei Lian, Shaohui Mei, Shun Zhang, and Mingyang Ma.
\newblock Benchmarking adversarial patch against aerial detection.
\newblock {\em IEEE Transactions on Geoscience and Remote Sensing}, 60:1--16, 2022.

\bibitem{lin2024out}
Tao Lin, Lijia Yu, Gaojie Jin, Renjue Li, Peng Wu, and Lijun Zhang.
\newblock Out-of-bounding-box triggers: A stealthy approach to cheat object detectors.
\newblock In {\em European Conference on Computer Vision}, pages 269--287. Springer, 2024.

\bibitem{wei2024physical}
Hui Wei, Hao Tang, Xuemei Jia, Zhixiang Wang, Hanxun Yu, Zhubo Li, Shin’ichi Satoh, Luc Van~Gool, and Zheng Wang.
\newblock Physical adversarial attack meets computer vision: A decade survey.
\newblock {\em IEEE Transactions on Pattern Analysis and Machine Intelligence}, 46(12):9797--9817, 2024.

\bibitem{wei2022visually}
Xingxing Wei, Bangzheng Pu, Jiefan Lu, and Baoyuan Wu.
\newblock Visually adversarial attacks and defenses in the physical world: A survey.
\newblock {\em arXiv preprint arXiv:2211.01671}, 2022.

\bibitem{guesmi2023physical}
Amira Guesmi, Muhammad~Abdullah Hanif, Bassem Ouni, and Muhammad Shafique.
\newblock Physical adversarial attacks for camera-based smart systems: Current trends, categorization, applications, research challenges, and future outlook.
\newblock {\em IEEE Access}, 11:109617--109668, 2023.

\bibitem{creswell2017latentpoison}
Antonia Creswell, Anil~A Bharath, and Biswa Sengupta.
\newblock Latentpoison-adversarial attacks on the latent space.
\newblock {\em arXiv preprint arXiv:1711.02879}, 2017.

\bibitem{jandial2019advgan}
Surgan Jandial, Puneet Mangla, Sakshi Varshney, and Vineeth Balasubramanian.
\newblock {AdvGAN++}: Harnessing latent layers for adversary generation.
\newblock In {\em Proceedings of the IEEE/CVF International Conference on Computer Vision (ICCV) Workshops}, pages 2045--2048, 2019.

\bibitem{inkawhich2019feature}
Nathan Inkawhich, Wei Wen, Hai~(Helen) Li, and Yiran Chen.
\newblock Feature space perturbations yield more transferable adversarial examples.
\newblock In {\em Proceedings of the IEEE/CVF Conference on Computer Vision and Pattern Recognition (CVPR)}, 2019.

\bibitem{zhou2019latent}
Xiaowei Zhou, Ivor~W Tsang, and Jie Yin.
\newblock Latent adversarial defence with boundary-guided generation.
\newblock {\em arXiv preprint arXiv:1907.07001}, 2019.

\bibitem{inkawhich2020transferable}
Nathan Inkawhich, Kevin Liang, Lawrence Carin, and Yiran Chen.
\newblock Transferable perturbations of deep feature distributions.
\newblock In {\em International Conference on Learning Representations}, 2020.

\bibitem{yu2021lafeat}
Yunrui Yu, Xitong Gao, and Cheng-Zhong Xu.
\newblock {LAFEAT}: Piercing through adversarial defenses with latent features.
\newblock In {\em Proceedings of the IEEE/CVF Conference on Computer Vision and Pattern Recognition (CVPR)}, pages 5735--5745, 2021.

\bibitem{wang2023generating}
Shuo Wang, Shangyu Chen, Tianle Chen, Surya Nepal, Carsten Rudolph, and Marthie Grobler.
\newblock Generating semantic adversarial examples via feature manipulation in latent space.
\newblock {\em IEEE Transactions on Neural Networks and Learning Systems}, 2023.

\bibitem{dunn2020semantic}
Isaac Dunn, Tom Melham, and Daniel Kroening.
\newblock Semantic adversarial perturbations using learnt representations.
\newblock {\em arXiv preprint arXiv:2001.11055}, 2020.

\bibitem{kumari2019harnessing}
Nupur Kumari, Mayank Singh, Abhishek Sinha, Harshitha Machiraju, Balaji Krishnamurthy, and Vineeth~N Balasubramanian.
\newblock Harnessing the vulnerability of latent layers in adversarially trained models.
\newblock In {\em Proceedings of the 28th International Joint Conference on Artificial Intelligence}, pages 2779--2785, 2019.

\bibitem{park2021reliably}
Geon~Yeong Park and Sang~Wan Lee.
\newblock Reliably fast adversarial training via latent adversarial perturbation.
\newblock In {\em Proceedings of the IEEE/CVF International Conference on Computer Vision (ICCV)}, pages 7758--7767, 2021.

\bibitem{upadhyay2021generating}
Ujjwal Upadhyay and Prerana Mukherjee.
\newblock Generating out of distribution adversarial attack using latent space poisoning.
\newblock {\em IEEE Signal Processing Letters}, 28:523--527, 2021.

\bibitem{schneider2022concept}
Johannes Schneider and Giovanni Apruzzese.
\newblock Concept-based adversarial attacks: Tricking humans and classifiers alike.
\newblock In {\em 2022 IEEE Security and Privacy Workshops (SPW)}, pages 66--72. IEEE, 2022.

\bibitem{wang2023semantic}
Chenan Wang, Jinhao Duan, Chaowei Xiao, Edward Kim, Matthew c~Stamm, and Kaidi Xu.
\newblock Semantic adversarial attacks via diffusion models.
\newblock In {\em 34th British Machine Vision Conference 2023, {BMVC} 2023, Aberdeen, UK, November 20-24, 2023}. BMVA, 2023.

\bibitem{zheng2023latent}
BoYang Zheng.
\newblock Latent magic: An investigation into adversarial examples crafted in the semantic latent space.
\newblock {\em arXiv preprint arXiv:2305.12906}, 2023.

\bibitem{clare2023generating}
Luana Clare and Jo{\~a}o Correia.
\newblock Generating adversarial examples through latent space exploration of generative adversarial networks.
\newblock In {\em Proceedings of the Companion Conference on Genetic and Evolutionary Computation}, pages 1760--1767, 2023.

\bibitem{wang2022semantic}
Xinyi Wang, Simon~Yusuf Enoch, and Dan~Dongseong Kim.
\newblock Semantic preserving adversarial attack generation with autoencoder and genetic algorithm.
\newblock In {\em GLOBECOM 2022 - 2022 IEEE Global Communications Conference}, pages 80--85, 2022.

\bibitem{shukla2023generating}
Nitish Shukla and Sudipta Banerjee.
\newblock Generating adversarial attacks in the latent space.
\newblock In {\em Proceedings of the IEEE/CVF International Conference on Computer Vision (ICCV) Workshops}, pages 730--739, 2023.

\bibitem{dale2024direct}
Ashley~S Dale and Lauren Christopher.
\newblock Direct adversarial latent estimation to evaluate decision boundary complexity in black box models.
\newblock {\em IEEE Transactions on Artificial Intelligence}, 2024.

\bibitem{li2024transcending}
Shuai Li, Xiaoyu Jiang, and Xiaoguang Ma.
\newblock Transcending adversarial perturbations: Manifold-aided adversarial examples with legitimate semantics.
\newblock {\em arXiv preprint arXiv:2402.03095}, 2024.

\bibitem{pan2024sca}
Zihao Pan, Weibin Wu, Yuhang Cao, and Zibin Zheng.
\newblock {SCA}: Highly efficient semantic-consistent unrestricted adversarial attack.
\newblock {\em CoRR}, 2024.

\bibitem{cao2022adversarial}
Yangjie Cao, Chenxi Zhu, Haobo Wang, and Yan Zhuang.
\newblock An adversarial attack algorithm based on edge-sketched feature from latent space.
\newblock In {\em 2022 2nd International Conference on Consumer Electronics and Computer Engineering (ICCECE)}, pages 723--728. IEEE, 2022.

\bibitem{casper2022robust}
Stephen Casper, Max Nadeau, Dylan Hadfield-Menell, and Gabriel Kreiman.
\newblock Robust feature-level adversaries are interpretability tools.
\newblock In S.~Koyejo, S.~Mohamed, A.~Agarwal, D.~Belgrave, K.~Cho, and A.~Oh, editors, {\em Advances in Neural Information Processing Systems}, volume~35, pages 33093--33106. Curran Associates, Inc., 2022.

\bibitem{guo2024generative}
Ziheng Guo, Zhongliang Guo, Oggie Arandelovic, and Andrea di~Falco.
\newblock Generative model for multiple-purpose inverse design and forward prediction of disordered waveguides in linear and nonlinear regimes.
\newblock In {\em Machine Learning in Photonics}, volume 13017, page 1301702. SPIE, 2024.

\bibitem{qian2025t2icount}
Yifei Qian, Zhongliang Guo, Bowen Deng, Chun~Tong Lei, Shuai Zhao, Chun~Pong Lau, Xiaopeng Hong, and Michael~P. Pound.
\newblock T2icount: Enhancing cross-modal understanding for zero-shot counting.
\newblock In {\em Proceedings of the Computer Vision and Pattern Recognition Conference (CVPR)}, pages 25336--25345, 2025.

\bibitem{lei2024instant}
Chun~Tong Lei, Hon~Ming Yam, Zhongliang Guo, Yifei Qian, and Chun~Pong Lau.
\newblock Instant adversarial purification with adversarial consistency distillation.
\newblock In {\em Proceedings of the Computer Vision and Pattern Recognition Conference (CVPR)}, pages 24331--24340, 2025.

\bibitem{qian2021improving}
Zhuang Qian, Shufei Zhang, Kaizhu Huang, Qiufeng Wang, Rui Zhang, and Xinping Yi.
\newblock Improving model robustness with latent distribution locally and globally.
\newblock {\em arXiv preprint arXiv:2107.04401}, 2021.

\bibitem{casper2021one}
Stephen Casper, Max Nadeau, and Gabriel Kreiman.
\newblock One thing to fool them all: Generating interpretable, universal, and physically-realizable adversarial features.
\newblock {\em Preprint}, 2021.

\bibitem{joshi2019semantic}
Ameya Joshi, Amitangshu Mukherjee, Soumik Sarkar, and Chinmay Hegde.
\newblock Semantic adversarial attacks: Parametric transformations that fool deep classifiers.
\newblock In {\em Proceedings of the IEEE/CVF International Conference on Computer Vision (ICCV)}, 2019.

\bibitem{jain2008biometric}
Anil~K. Jain, Karthik Nandakumar, and Abhishek Nagar.
\newblock Biometric template security.
\newblock {\em EURASIP J. Adv. Signal Process.}, 2008, 2008.

\bibitem{ratha2006cancelable}
Nalini Ratha, Jonathan Connell, Ruud~M Bolle, and Sharat Chikkerur.
\newblock Cancelable biometrics: A case study in fingerprints.
\newblock In {\em 18th International Conference on Pattern Recognition (ICPR'06)}, volume~4, pages 370--373. IEEE, 2006.

\bibitem{ratha2007generating}
Nalini~K Ratha, Sharat Chikkerur, Jonathan~H Connell, and Ruud~M Bolle.
\newblock Generating cancelable fingerprint templates.
\newblock {\em IEEE Transactions on Pattern Analysis and Machine Intelligence}, 29(4):561--572, 2007.

\bibitem{boult2007revocable}
Terrance~E Boult, Walter~J Scheirer, and Robert Woodworth.
\newblock Revocable fingerprint biotokens: Accuracy and security analysis.
\newblock In {\em 2007 IEEE Conference on Computer Vision and Pattern Recognition (CVPR)}, pages 1--8. IEEE, 2007.

\bibitem{jin2010revocable}
Zhe Jin, Andrew Beng~Jin Teoh, Thian~Song Ong, and Connie Tee.
\newblock A revocable fingerprint template for security and privacy preserving.
\newblock {\em KSII Transactions on Internet and Information Systems (TIIS)}, 4(6):1327--1342, 2010.

\bibitem{biggio2015adversarial}
Battista Biggio, Paolo Russu, Luca Didaci, Fabio Roli, et~al.
\newblock Adversarial biometric recognition: A review on biometric system security from the adversarial machine-learning perspective.
\newblock {\em IEEE Signal Processing Magazine}, 32(5):31--41, 2015.

\bibitem{guo2024white}
Zhongliang Guo, Weiye Li, Yifei Qian, Ognjen Arandjelovic, and Lei Fang.
\newblock A white-box false positive adversarial attack method on contrastive loss based offline handwritten signature verification models.
\newblock In {\em International Conference on Artificial Intelligence and Statistics}, pages 901--909. PMLR, 2024.

\bibitem{lovisotto2020biometric}
Giulio Lovisotto, Simon Eberz, and Ivan Martinovic.
\newblock Biometric backdoors: A poisoning attack against unsupervised template updating.
\newblock In {\em 2020 IEEE European Symposium on Security and Privacy (EuroS\&P)}, pages 184--197. IEEE, 2020.

\bibitem{won2021generative}
Hee won Kwon, Jea-Won Nam, Joongheon Kim, and Youn~Kyu Lee.
\newblock Generative adversarial attacks on fingerprint recognition systems.
\newblock In {\em 2021 International Conference on Information Networking (ICOIN)}, pages 483--485. IEEE, 2021.

\bibitem{marrone2019adversarial}
Stefano Marrone and Carlo Sansone.
\newblock Adversarial perturbations against fingerprint based authentication systems.
\newblock In {\em 2019 International Conference on Biometrics (ICB)}, pages 1--6. IEEE, 2019.

\bibitem{hafemann2019characterizing}
Luiz~G Hafemann, Robert Sabourin, and Luiz~S Oliveira.
\newblock Characterizing and evaluating adversarial examples for offline handwritten signature verification.
\newblock {\em IEEE Transactions on Information Forensics and Security}, 14(8):2153--2166, 2019.

\bibitem{liu2024mmad}
Xinxin Liu, Zhongliang Guo, Siyuan Huang, and Chun~Pong Lau.
\newblock Mmad-purify: A precision-optimized framework for efficient and scalable multi-modal attacks.
\newblock {\em arXiv preprint arXiv:2410.14089}, 2024.

\bibitem{zhao2024weak}
Shuai Zhao, Leilei Gan, Zhongliang Guo, Xiaobao Wu, Luwei Xiao, Xiaoyu Xu, Cong-Duy Nguyen, and Luu~Anh Tuan.
\newblock Weak-to-strong backdoor attack for large language models.
\newblock {\em arXiv preprint arXiv:2409.17946}, 2024.

\bibitem{li2025fedkd}
Yuqi Li, Xingyou Lin, Kai Zhang, Chuanguang Yang, Zhongliang Guo, Jianping Gou, and Yanli Li.
\newblock Fedkd-hybrid: Federated hybrid knowledge distillation for lithography hotspot detection.
\newblock {\em arXiv preprint arXiv:2501.04066}, 2025.

\bibitem{lei2025towards}
Chun~Tong Lei, Zhongliang Guo, Hon~Chung Lee, Minh~Quoc Duong, and Chun~Pong Lau.
\newblock Towards more transferable adversarial attack in black-box manner.
\newblock {\em arXiv preprint arXiv:2505.18097}, 2025.

\bibitem{yam2025my}
Hon~Ming Yam, Zhongliang Guo, and Chun~Pong Lau.
\newblock My face is mine, not yours: Facial protection against diffusion model face swapping.
\newblock {\em arXiv preprint arXiv:2505.15336}, 2025.

\bibitem{zhao2025affective}
Shuai Zhao, Yulin Zhang, Luwei Xiao, Xinyi Wu, Yanhao Jia, Zhongliang Guo, Xiaobao Wu, Cong-Duy Nguyen, Guoming Zhang, and Anh~Tuan Luu.
\newblock Affective-roptester: Capability and bias analysis of llms in predicting retinopathy of prematurity.
\newblock {\em arXiv preprint arXiv:2507.05816}, 2025.

\bibitem{poh2012critical}
Norman Poh, Ajita Rattani, and Fabio Roli.
\newblock Critical analysis of adaptive biometric systems.
\newblock {\em IET Biometrics}, 1(4):179--187, 2012.

\bibitem{park2024comprehensive}
Seong~Hee Park, Soo-Hyun Lee, Min~Young Lim, Pyo~Min Hong, and Youn~Kyu Lee.
\newblock A comprehensive risk analysis method for adversarial attacks on biometric authentication systems.
\newblock {\em IEEE Access}, 2024.

\bibitem{lavens2023mitigating}
Emma Lavens, Davy Preuveneers, and Wouter Joosen.
\newblock Mitigating undesired interactions between liveness detection components in biometric authentication.
\newblock In {\em Proceedings of the 18th International Conference on Availability, Reliability and Security}, pages 1--8, 2023.

\bibitem{islam2017increasing}
Tasmina Islam.
\newblock {\em Increasing Reliability and Security in Handwritten Signature Biometrics}.
\newblock University of Kent (United Kingdom), 2017.

\bibitem{walia2019adaptive}
Gurjit~Singh Walia, Gaurav Jain, Nipun Bansal, and Kuldeep Singh.
\newblock Adaptive weighted graph approach to generate multimodal cancelable biometric templates.
\newblock {\em IEEE Transactions on Information Forensics and Security}, 15:1945--1958, 2019.

\bibitem{alghamdi2024enhancing}
Shaima~M Alghamdi, Salma~Kammoun Jarraya, and Faris Kateb.
\newblock Enhancing security in multimodal biometric fusion: Analyzing adversarial attacks.
\newblock {\em IEEE Access}, 2024.

\bibitem{lee2024adversarial}
MyeongHoe Lee, JunHo Yoon, and Chang Choi.
\newblock Adversarial attack vulnerability for multi-biometric authentication system.
\newblock {\em Expert Systems}, 41(10):e13655, 2024.

\bibitem{advdm}
Chumeng Liang, Xiaoyu Wu, Yang Hua, Jiaru Zhang, Yiming Xue, Tao Song, Zhengui Xue, Ruhui Ma, and Haibing Guan.
\newblock Adversarial example does good: Preventing painting imitation from diffusion models via adversarial examples.
\newblock In {\em International Conference on Machine Learning}, pages 20763--20786. PMLR, 2023.

\bibitem{mist}
Chumeng Liang and Xiaoyu Wu.
\newblock Mist: Towards improved adversarial examples for diffusion models.
\newblock {\em arXiv preprint arXiv:2305.12683}, 2023.

\bibitem{zhao2024unlearnable}
Zhengyue Zhao, Jinhao Duan, Xing Hu, Kaidi Xu, Chenan Wang, Rui Zhang, Zidong Du, Qi~Guo, and Yunji Chen.
\newblock Unlearnable examples for diffusion models: Protect data from unauthorized exploitation.
\newblock In {\em ICLR 2024 Workshop on Reliable and Responsible Foundation Models}, 2024.

\bibitem{ruiz2023dreambooth}
Nataniel Ruiz, Yuanzhen Li, Varun Jampani, Yael Pritch, Michael Rubinstein, and Kfir Aberman.
\newblock {DreamBooth}: Fine tuning text-to-image diffusion models for subject-driven generation.
\newblock In {\em Proceedings of the IEEE/CVF Conference on Computer Vision and Pattern Recognition (CVPR)}, pages 22500--22510, 2023.

\bibitem{le2023anti}
Thanh Van~Le, Hao Phung, Thuan~Hoang Nguyen, Quan Dao, Ngoc~N. Tran, and Anh Tran.
\newblock {Anti-DreamBooth}: Protecting users from personalized text-to-image synthesis.
\newblock In {\em Proceedings of the IEEE/CVF International Conference on Computer Vision (ICCV)}, pages 2116--2127, 2023.

\bibitem{liu2024metacloak}
Yixin Liu, Chenrui Fan, Yutong Dai, Xun Chen, Pan Zhou, and Lichao Sun.
\newblock {MetaCloak}: Preventing unauthorized subject-driven text-to-image diffusion-based synthesis via meta-learning.
\newblock In {\em Proceedings of the IEEE/CVF Conference on Computer Vision and Pattern Recognition (CVPR)}, pages 24219--24228, 2024.

\bibitem{wang2024simac}
Feifei Wang, Zhentao Tan, Tianyi Wei, Yue Wu, and Qidong Huang.
\newblock {SimAC}: A simple anti-customization method for protecting face privacy against text-to-image synthesis of diffusion models.
\newblock In {\em Proceedings of the IEEE/CVF Conference on Computer Vision and Pattern Recognition (CVPR)}, pages 12047--12056, 2024.

\bibitem{guo2024real}
Hanzhong Guo, Shen Nie, Chao Du, Tianyu Pang, Hao Sun, and Chongxuan Li.
\newblock Real-time identity defenses against malicious personalization of diffusion models.
\newblock {\em arXiv preprint arXiv:2412.09844}, 2024.

\bibitem{wan2024prompt}
Cong Wan, Yuhang He, Xiang Song, and Yihong Gong.
\newblock Prompt-agnostic adversarial perturbation for customized diffusion models.
\newblock {\em Advances in Neural Information Processing Systems}, 37:136576--136619, 2024.

\bibitem{sds}
Haotian Xue, Chumeng Liang, Xiaoyu Wu, and Yongxin Chen.
\newblock Toward effective protection against diffusion-based mimicry through score distillation.
\newblock In {\em International Conference on Learning Representations}, 2024.

\bibitem{wu2023towards}
Ruijia Wu, Yuhang Wang, Huafeng Shi, Zhipeng Yu, Yichao Wu, and Ding Liang.
\newblock Towards prompt-robust face privacy protection via adversarial decoupling augmentation framework.
\newblock {\em arXiv preprint arXiv:2305.03980}, 2023.

\bibitem{ozden2024optimization}
Tarik~Can Ozden, Ozgur Kara, Oguzhan Akcin, Kerem Zaman, Shashank Srivastava, Sandeep~P Chinchali, and James~M Rehg.
\newblock Optimization-free image immunization against diffusion-based editing.
\newblock {\em arXiv preprint arXiv:2411.17957}, 2024.

\bibitem{zhu2024watermark}
Peifei Zhu, Tsubasa Takahashi, and Hirokatsu Kataoka.
\newblock Watermark-embedded adversarial examples for copyright protection against diffusion models.
\newblock In {\em Proceedings of the IEEE/CVF Conference on Computer Vision and Pattern Recognition (CVPR)}, pages 24420--24430, 2024.

\bibitem{tan2023somewhat}
Mingtian Tan, Tianhao Wang, and Somesh Jha.
\newblock A somewhat robust image watermark against diffusion-based editing models.
\newblock {\em arXiv preprint arXiv:2311.13713}, 2023.

\bibitem{cao2023impress}
Bochuan Cao, Changjiang Li, Ting Wang, Jinyuan Jia, Bo~Li, and Jinghui Chen.
\newblock Impress: Evaluating the resilience of imperceptible perturbations against unauthorized data usage in diffusion-based generative ai.
\newblock {\em Advances in Neural Information Processing Systems}, 36:10657--10677, 2023.

\bibitem{zhao2024can}
Zhengyue Zhao, Jinhao Duan, Kaidi Xu, Chenan Wang, Rui Zhang, Zidong Du, Qi~Guo, and Xing Hu.
\newblock Can protective perturbation safeguard personal data from being exploited by stable diffusion?
\newblock In {\em Proceedings of the IEEE/CVF Conference on Computer Vision and Pattern Recognition (CVPR)}, pages 24398--24407, 2024.

\bibitem{an2024rethinking}
Shengwei An, Lu~Yan, Siyuan Cheng, Guangyu Shen, Kaiyuan Zhang, Qiuling Xu, Guanhong Tao, and Xiangyu Zhang.
\newblock Rethinking the invisible protection against unauthorized image usage in stable diffusion.
\newblock In {\em 33rd USENIX Security Symposium (USENIX Security 24)}, pages 3621--3638, 2024.

\bibitem{choi2025diffusionguard}
June~Suk Choi, Kyungmin Lee, Jongheon Jeong, Saining Xie, Jinwoo Shin, and Kimin Lee.
\newblock {DiffusionGuard}: A robust defense against malicious diffusion-based image editing.
\newblock In {\em International Conference on Learning Representations}, 2025.

\bibitem{yang2024ddap}
Jing Yang, Runping Xi, Yingxin Lai, Xun Lin, and Zitong Yu.
\newblock {DDAP}: Dual-domain anti-personalization against text-to-image diffusion models.
\newblock In {\em 2024 IEEE International Joint Conference on Biometrics (IJCB)}, pages 1--10. IEEE, 2024.

\bibitem{mi2024visual}
Xiaoyue Mi, Fan Tang, Juan Cao, Peng Li, and Yang Liu.
\newblock Visual-friendly concept protection via selective adversarial perturbations.
\newblock {\em CoRR}, 2024.

\bibitem{mopuri2017fast}
Konda~Reddy Mopuri, Utsav Garg, and R~Venkatesh Babu.
\newblock Fast feature fool: A data independent approach to universal adversarial perturbations.
\newblock In {\em Proceedings of the British Machine Vision Conference ({BMVC})}, 2017.

\bibitem{lu2022domaininvariant}
Wang Lu, Jindong Wang, Haoliang Li, Yiqiang Chen, and Xing Xie.
\newblock Domain-invariant feature exploration for domain generalization.
\newblock {\em Transactions on Machine Learning Research}, 2022.

\bibitem{radford2021learning}
Alec Radford, Jong~Wook Kim, Chris Hallacy, Aditya Ramesh, Gabriel Goh, Sandhini Agarwal, Girish Sastry, Amanda Askell, Pamela Mishkin, Jack Clark, et~al.
\newblock Learning transferable visual models from natural language supervision.
\newblock In {\em International Conference on Machine Learning}, pages 8748--8763. PmLR, 2021.

\bibitem{oquabdinov2}
Maxime Oquab, Timoth{\'e}e Darcet, Th{\'e}o Moutakanni, Huy~V. Vo, Marc Szafraniec, Vasil Khalidov, Pierre Fernandez, Daniel HAZIZA, Francisco Massa, Alaaeldin El-Nouby, Mido Assran, Nicolas Ballas, Wojciech Galuba, Russell Howes, Po-Yao Huang, Shang-Wen Li, Ishan Misra, Michael Rabbat, Vasu Sharma, Gabriel Synnaeve, Hu~Xu, Herve Jegou, Julien Mairal, Patrick Labatut, Armand Joulin, and Piotr Bojanowski.
\newblock {DINO}v2: Learning robust visual features without supervision.
\newblock {\em Transactions on Machine Learning Research}, 2024.
\newblock Featured Certification.

\bibitem{bai2024sequential}
Yutong Bai, Xinyang Geng, Karttikeya Mangalam, Amir Bar, Alan~L Yuille, Trevor Darrell, Jitendra Malik, and Alexei~A Efros.
\newblock Sequential modeling enables scalable learning for large vision models.
\newblock In {\em Proceedings of the IEEE/CVF Conference on Computer Vision and Pattern Recognition (CVPR)}, pages 22861--22872, 2024.

\bibitem{biamonte2017quantum}
Jacob Biamonte, Peter Wittek, Nicola Pancotti, Patrick Rebentrost, Nathan Wiebe, and Seth Lloyd.
\newblock Quantum machine learning.
\newblock {\em Nature}, 549(7671):195--202, 2017.

\end{thebibliography}

\end{document}